%% file: main.tex
\documentclass[10pt,letterpaper]{lab}

\usepackage[utf8]{inputenc}
\usepackage[T1]{fontenc}
\usepackage{charter}
\input{preamble}
\hypersetup{pdftitle={Gravis: Embodied-Native Pretraining for Generalizable World Action Models}}

\title{Gravis: Embodied-Native Pretraining for Generalizable World Action Models}
\author[1,2]{Shuhe Huang}
\author[1,2]{Hengkai Tan}
\author[2]{Gang Yue}
\author[1,2]{Chendong Xiang}
\author[1,2]{Zeyuan Wang}
\author[2]{Xiaolong Liu}
\author[2,3]{Rongxu Cui}
\author[2]{Xiao Liu}
\author[2]{Jiangfeng Liu}
\author[1,2]{Zihao Sun}
\author[1,2]{Chengwen Zhang}
\author[1,2]{Ruowen Zhao}
\author[1,2]{Jun Zhu}
\affiliation[1]{Tsinghua University}
\affiliation[2]{GensPI Technology}
\affiliation[3]{Beihang University}

\begin{document}
\maketitle

\begin{tcolorbox}[
  enhanced,
  breakable,
  colback=labblue!10,
  colframe=labblue!35,
  boxrule=0.5pt,
  arc=6pt,
  left=10pt,
  right=10pt,
  top=2pt,
  bottom=2pt
]
\input{sec/0_abstract}
\end{tcolorbox}

\noindent\begin{minipage}{\textwidth}
\centering
\includegraphics[width=\textwidth]{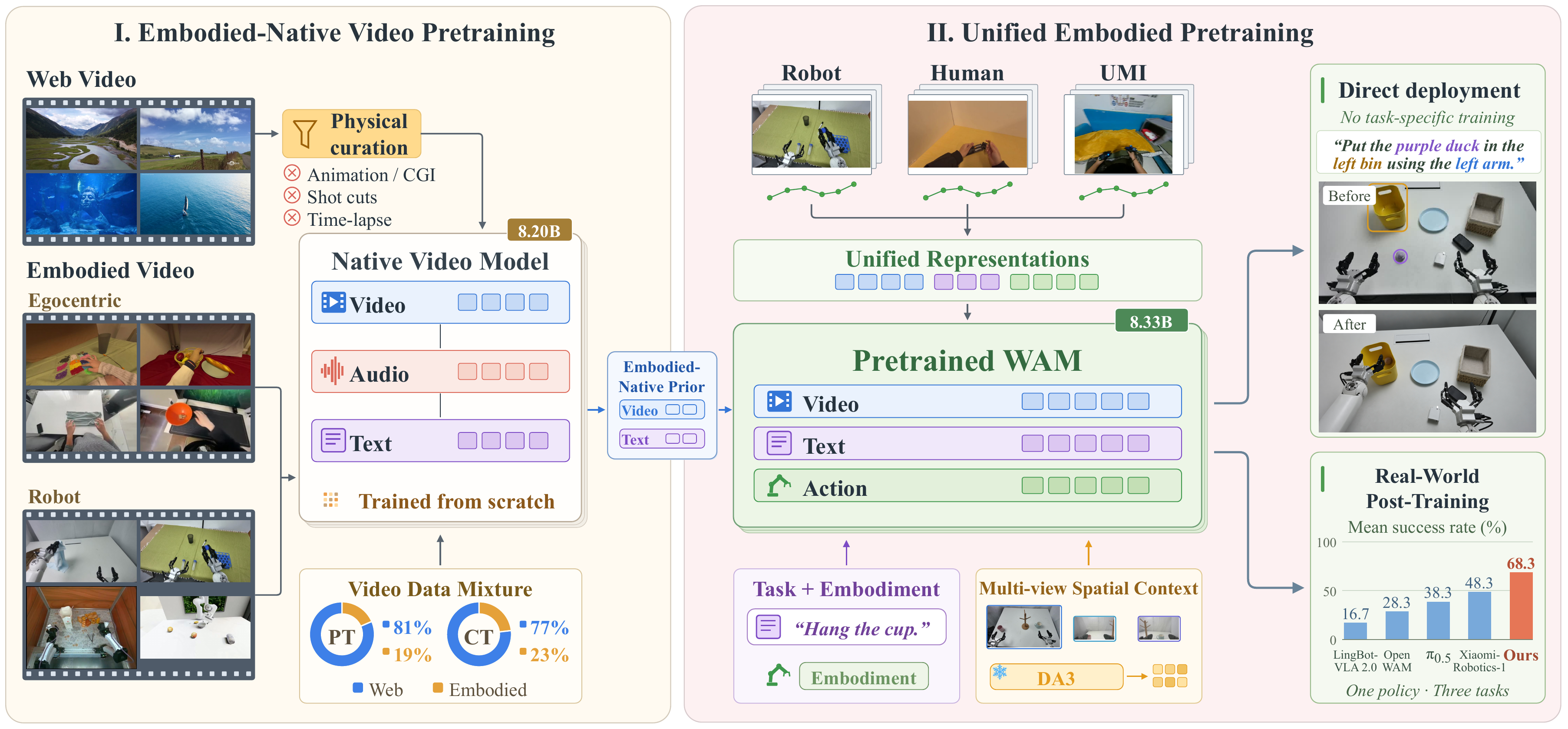}
\captionsetup{hypcap=false}
\captionof{figure}{\textbf{Overview of Gravis: two-stage embodied-native pretraining.} Gravis-Video learns visual interaction priors from curated Internet videos and embodied demonstrations. Building on this foundation, Gravis-WAM aligns heterogeneous robot, human, and UMI data through unified action representations, native multi-view modeling, and embodiment conditioning, with depth-derived spatial context.}
\label{fig:pipeline}
\end{minipage}

\clearpage

\input{sec/1_intro}

\begin{figure}[!t]
\centering
\includegraphics[width=\textwidth]{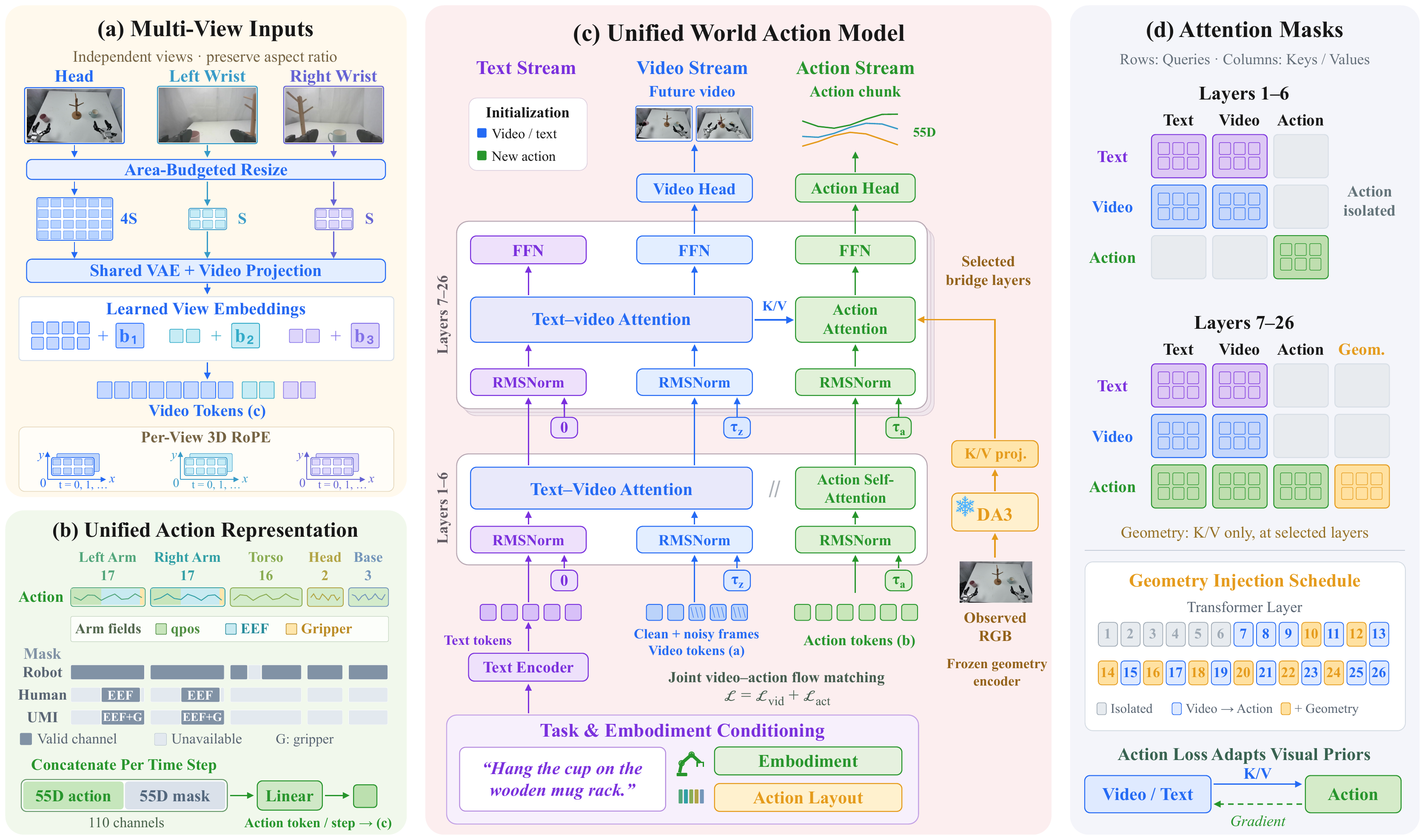}
\caption{\textbf{Gravis-WAM architecture.} (a) Area-budgeted resizing, independent encoding, view embeddings, and per-view 3D RoPE accommodate heterogeneous cameras. (b) Actions share a masked 55D representation. (c) Independent noise levels control video/action generation; frozen DA3 encodes observed geometry. (d) Actions are processed separately in layers 1--6, then attend to video/text in layers 7--26, with geometry injected at selected layers. Video/text attention excludes actions, while action-loss gradients adapt visual priors.}
\label{fig:architecture}
\end{figure}

\input{sec/2_related}

\input{sec/3_method}
\input{sec/4_experiments}
\input{sec/6_conclusion}

\FloatBarrier
\bibliographystyle{unsrtnat}
\bibliography{main}

\clearpage
\noindent{\Large\bfseries\color{labblue} Appendix}\par\medskip
\input{sec/supplementary}

\end{document}

%% file: preamble.tex
\usepackage{graphicx}
\usepackage{amsmath,amssymb}
\usepackage{booktabs,multirow,array,tabularx}
\usepackage{xcolor}
\usepackage{microtype}
\usepackage{needspace}
\usepackage{enumitem}
\definecolor{paperblue}{HTML}{376EAC}
\definecolor{paperteal}{HTML}{22877F}
\definecolor{papergray}{HTML}{5F6C7D}
\definecolor{paperpale}{HTML}{EDF3FA}
\definecolor{draftamber}{HTML}{996320}

\newcommand{\piHalf}{$\pi_{0.5}$}
\newcolumntype{Y}{>{\raggedright\arraybackslash}X}
\setlist[itemize]{leftmargin=*,topsep=3pt,itemsep=2pt,parsep=0pt}
\renewcommand{\arraystretch}{1.05}
\graphicspath{{figures/}}

\renewcommand{\topfraction}{.9}
\renewcommand{\bottomfraction}{.85}
\renewcommand{\textfraction}{.08}
\renewcommand{\floatpagefraction}{.82}
\AtBeginDocument{\raggedbottom}

%% file: sec/0_abstract.tex
\abstract{
Building embodied-native foundation models requires learning both physical dynamics and executable behavior.
However, general-purpose video pretraining prioritizes content generation over embodied interaction, while heterogeneous embodiments, action spaces, and camera configurations complicate joint embodied learning.
To address these challenges, we introduce \textbf{Gravis}, a two-stage embodied-native pretraining framework that learns visual interaction priors from curated real-world videos and grounds them in executable actions through unified learning from aligned heterogeneous embodied experience.
First, we train \textbf{Gravis-Video} from scratch on real-world videos curated for physical continuity and enriched with robot manipulation and egocentric human interactions.
We then construct \textbf{Gravis-WAM} and align heterogeneous human and robot data through native multi-view modeling, unified action representations, and explicit embodiment conditioning, retaining the spatial and embodiment-specific context needed for embodied understanding and behavior learning.
\textbf{Gravis-Video} achieves the highest mean embodiment consistency on our benchmark and outperforms the evaluated open-weight generators in physical and embodiment consistency, supporting the acquisition of strong embodied priors through embodied-native pretraining.
\textbf{Gravis-WAM} demonstrates strong simulated control, direct real-world deployment, and effective real-world multi-task adaptation.
Together, these results support learning embodied visual priors and aligning heterogeneous experience as complementary foundations for generalizable robot behavior.
}

%% file: sec/1_intro.tex
\section{Introduction}
\label{sec:intro}
Recent World Action Models (WAMs) connect video prediction with action generation to learn robot behavior from diverse visual and embodied experience~\cite{motus,dreamzero,lingbotva,motubrain,lingbotva2,uwm,cosmospolicy,fastwam}.
Their progress motivates \emph{embodied-native pretraining}: learning visual priors and action representations specifically suited to physical interaction.

This objective calls for reconsidering both the video foundations and the embodied data representations used in WAM pretraining.
General-purpose video pretraining primarily serves content generation, using corpora that include animation and discontinuous sequences~\cite{seedance20,vidu}.
Strong perceptual quality therefore does not guarantee suitable priors for embodied interaction, motivating pretraining that prioritizes physical continuity and embodiment consistency.
Building WAMs on these foundations further requires aligning data representations across heterogeneous camera configurations and action conventions.
Image tiling scales poorly, may disrupt pretrained spatial structure, and provides limited cross-view geometry; inconsistent action semantics can introduce conflicting supervision.
Joint pretraining thus requires reducing dataset mismatches while retaining spatial and embodiment-specific context.

We introduce \textbf{Gravis}, a two-stage embodied-native pretraining framework.
First, we pretrain \textbf{Gravis-Video} from scratch on real-world videos filtered for real-world physical continuity and enriched with robot manipulation and egocentric human interactions.
General videos provide broad commonsense coverage, while embodied footage and higher-resolution continued training emphasize fine-grained physical interactions.
Second, we construct \textbf{Gravis-WAM} and align heterogeneous embodied data for joint video--action pretraining.
Independent area-based resizing, per-view positional encoding, and view embeddings preserve local visual structure while enabling multi-view learning.
Depth-derived features~\cite{da3} add spatial context; shared masked actions, camera-relative end-effector motions, and embodiment conditioning reconcile action conventions.
Video-to-action attention connects action prediction to pretrained visual features.
Together, these designs reduce representational conflicts across data sources and enable shared interaction knowledge to be learned from heterogeneous embodiments.

We evaluate Gravis from embodied video generation to closed-loop robot control.
Gravis-Video achieves the highest mean embodiment consistency, with significant physical- and embodiment-consistency gains over evaluated open-weight generators.
Gravis-WAM performs strongly on RoboTwin~\cite{robotwin} and EBench~\cite{ebench}, reaches 80.42\% success through direct real-world deployment, and improves three-task post-training success by 20.0 percentage points over the strongest baseline.
Embodied pretraining further improves Clean2Random success by 14.48 points, demonstrating transfer beyond downstream training conditions.

Our contributions are fourfold:
\begin{itemize}
\item We introduce \textbf{Gravis}, a two-stage framework connecting embodied-native video pretraining with heterogeneous embodied pretraining.
\item We develop \textbf{Gravis-Video} through training from scratch on curated videos with progressive resolution, emphasizing physical continuity and embodiment consistency.
\item We develop \textbf{Gravis-WAM} to align heterogeneous training data through native multi-view modeling, depth-derived spatial features, and embodiment-conditioned shared actions.
\item We demonstrate that \textbf{Gravis} supports strong embodied video generation and simulated manipulation, while enabling direct deployment of the pretrained policy and effective multi-task adaptation on real robots.
\end{itemize}

%% file: sec/2_related.tex
\section{Related Work}
\label{sec:related}
\paragraph{Video Generation and Physical Understanding.}
Large-scale video models, including Wan, Seedance, Kling, and Vidu~\cite{wan,seedance,vidu}, advance conditional generation and visual dynamics modeling.
Yet perceptual realism does not ensure physical consistency or generalization to unseen dynamics~\cite{videophy,physicsiq,physicalgeneralization}.
% Embodied motivation follows directly from the limitations above.
Gravis-Video trains from scratch on curated real-world videos enriched with robot and egocentric human interactions, prioritizing physical continuity and embodiment consistency while retaining broad visual coverage.

\paragraph{World Action Models.}
Early approaches infer actions from predicted videos through inverse dynamics~\cite{vidar,mimicvideo,robodreamer,vidarc2025}, as in UniPi~\cite{unipi} and Video Prediction Policy~\cite{vpp}.
Recent WAMs jointly model future observations and actions, integrating visual dynamics with policy learning~\cite{uwm,motus,dreamzero,lingbotva,cosmospolicy,openwamexp,motubrain}.

WAM pretraining strategies differ in how they acquire and transfer visual priors.
Motus~\cite{motus} adapts a pretrained video generator through embodied-video and latent-action training before learning robot actions.
DreamZero~\cite{dreamzero} jointly learns video and action prediction from a pretrained video backbone, while Unified World Models~\cite{uwm} incorporates action-free videos through coupled diffusion.
LingBot-VA~2.0~\cite{lingbotva2} instead jointly learns generation and control from scratch.
Complementary efforts explore cross-embodiment policy learning~\cite{openx,octo,openvla} and human--robot data alignment~\cite{umi,egomimic}.
Gravis adopts two-stage embodied-native pretraining: it first learns visual interaction priors from scratch using physically curated real-world videos and embodied demonstrations, then jointly pretrains video--action prediction on aligned heterogeneous human and robot data through native multi-view modeling, geometry conditioning, unified action representations, and explicit embodiment conditioning.

%% file: sec/3_method.tex
\section{Method}
\label{sec:method}
\paragraph{Overview.}
Gravis models language-conditioned manipulation through joint generation of future videos $\mathbf{v}_t^{+}$ and actions $\mathbf{a}_t$:
\begin{equation}
p_\theta(\mathbf{v}_t^{+},\mathbf{a}_t\mid\mathbf{o}_t,\ell,e),
\label{eq:jointdistribution}
\end{equation}
conditioned on current multi-view observations $\mathbf{o}_t$, instruction $\ell$, and embodiment descriptor $e$.
We first pretrain Gravis-Video from scratch to acquire visual interaction priors (Sec.~\ref{sec:video}), then construct Gravis-WAM and align heterogeneous embodied data through native multi-view modeling, geometry conditioning, and shared actions (Secs.~\ref{sec:transfer}--\ref{sec:embodied}).

\subsection{Gravis-Video: Embodied-Native Pretraining}
\label{sec:video}
\label{sec:data}
\paragraph{Progressive pretraining and data composition.}
Gravis-Video undergoes pretraining (PT) and continued training (CT).
We collect an Internet video corpus and curate it for continuous real-world dynamics, excluding animation, simulated or rendered content, shot transitions, and time-lapse footage.
The retained videos are mixed with egocentric human manipulation, robot demonstrations, and simulated interactions.
General/embodied videos constitute 81\%/19\% in PT and 77\%/23\% in CT; egocentric videos account for 59\% and 68\% of the respective embodied subsets.
CT increases the target frame long side from 512 to 896 pixels to retain finer manipulation details.
Both stages use the Wan2.2 VAE~\cite{wan22}.

\paragraph{Three-stream model architecture.}
The 26-block transformer comprises video, audio, and text streams with 8.20B total parameters; the video-generation pathway, including text processing and shared time conditioning, contains 5.45B.
Modality-specific projections embed video/audio latents and T5Gemma~\cite{t5gemma} text features into a shared hidden space.
Each stream has separate attention projections and ReGLU feed-forward networks.
Joint attention uses RMS-normalized queries and keys, with 3D RoPE for video and 1D RoPE for audio; separate heads predict video/audio flow velocities.
The flow-matching timestep $\tau$ is encoded by a sinusoidal embedding and a shared MLP.
This embedding modulates feature scales and residual gates in each modality's transformer blocks, and feature scales in the video and audio output heads.
Generated tokens use the sampled $\tau$, while observed frames and text use $\tau=0$.
Synchronized audio supplies complementary supervision for visual dynamics learning~\cite{cosmos3}.

\begin{figure}[!htbp]
\centering
\includegraphics[width=\textwidth]{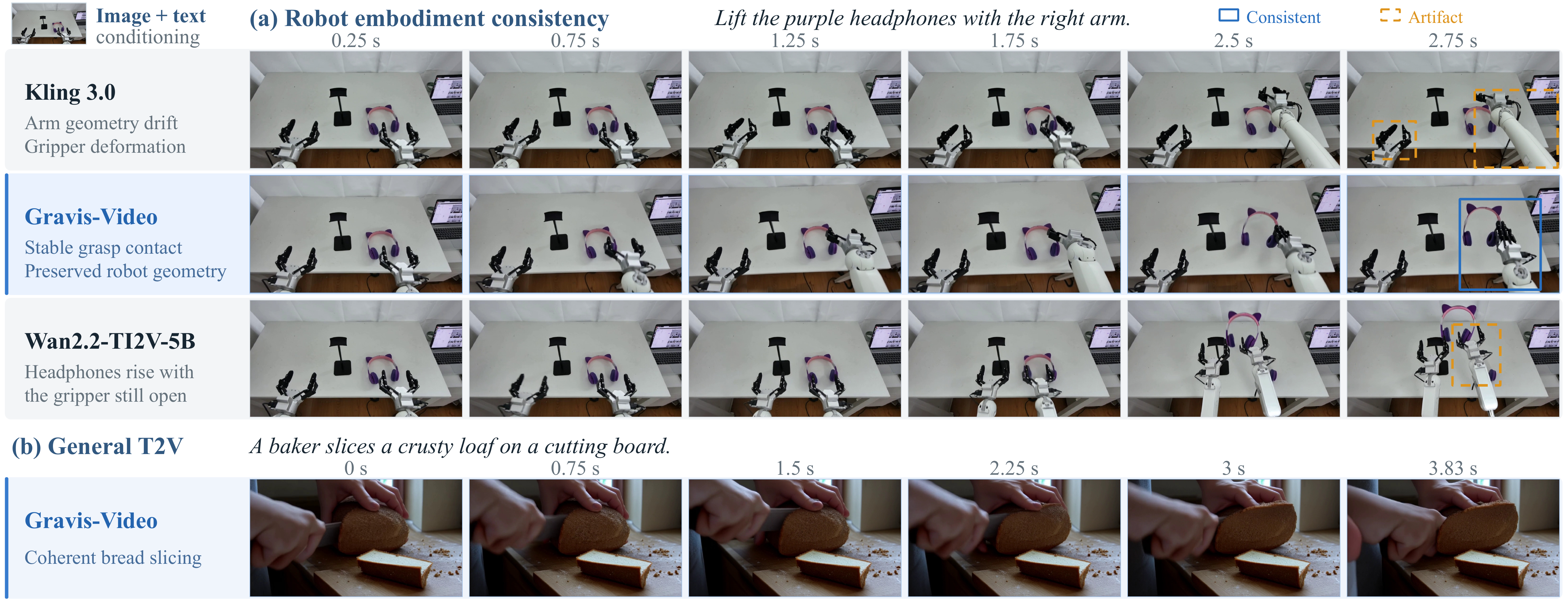}
\caption{\textbf{Qualitative video generation.} (a) Robot TI2V from the same initial image at matched timestamps. Gravis-Video maintains robot geometry and grasp contact (blue boxes); baselines show distortions or unsupported lifting (amber boxes). (b) Text-only generation demonstrates capabilities beyond robotic scenes.}
\label{fig:video_generation}
\end{figure}

\begin{figure}[!t]
\centering
\includegraphics[width=\textwidth]{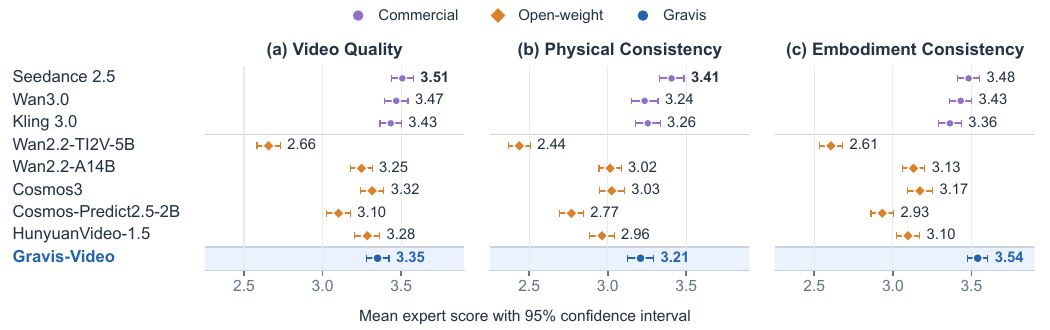}
\caption{\textbf{Expert video evaluation.} Example-averaged 1--5 scores on 200 TI2V cases; bars denote 95\% case-bootstrap CIs conditional on the 14-expert panel. Although its VQ trails leading closed-source models, Gravis-Video achieves the highest mean EC among evaluated generators, suggesting that stronger general-purpose video generation does not necessarily imply better embodied priors.}
\label{fig:video_scores}
\end{figure}

\paragraph{Training objective.}
We use flow matching~\cite{flowmatching} with an 8:2 T2V/TI2V ratio.
For clean video latents $\mathbf{z}_i$ and Gaussian noise $\boldsymbol{\epsilon}_i$,
\begin{equation}
\mathbf{z}_{i,\tau_i}=(1-\tau_i)\mathbf{z}_i+\tau_i\boldsymbol{\epsilon}_i,\qquad
\mathbf{u}_{\mathrm{vid},i}=\boldsymbol{\epsilon}_i-\mathbf{z}_i.
\label{eq:path}
\end{equation}
All predicted frames within clip $i$ share the sampled timestep $\tau_i$.
Conditioning frames remain clean with $\tau=0$.
The per-clip video loss is defined as
\begin{equation}
\mathcal{L}_{\mathrm{vid},i}
=\frac{\left\|\mathbf{m}_{\mathrm{vid},i}\odot
(\widehat{\mathbf{u}}_{\mathrm{vid},i}-\mathbf{u}_{\mathrm{vid},i})\right\|_2^2}
{n_{\mathrm{vid},i}},
\label{eq:video}
\end{equation}
where $\widehat{\mathbf{u}}$ is the predicted velocity, $\mathbf{m}$ selects predicted elements, $\odot$ denotes element-wise multiplication, and $n_{\mathrm{vid},i}$ counts all latent elements, including conditioning frames.
Per-clip reduction prevents longer or higher-resolution videos from receiving greater weight solely through token counts.
Subscripts $\mathrm{vid}$ and $\mathrm{aud}$ denote video and audio; the analogous audio loss uses the same $\tau_i$ and excludes padding.
For a batch of $b$ samples,
\begin{equation}
\mathcal{L}_{\mathrm{pre}}
=\frac{1}{b}\sum_{i=1}^{b}
\begin{cases}
2\mathcal{L}_{\mathrm{vid},i}, & \text{without audio},\\
1.5(\mathcal{L}_{\mathrm{vid},i}+\mathcal{L}_{\mathrm{aud},i}), & \text{with audio}.
\end{cases}
\label{eq:pretraining}
\end{equation}

\subsection{Gravis-WAM: Aligning Heterogeneous Embodied Data}
\label{sec:transfer}
\paragraph{Video-to-action transfer.}
Gravis-WAM inherits the video and text pathways and time conditioning, replacing audio with a newly initialized action stream.
It contains 8.33B trainable parameters, including view embeddings and geometry projections but excluding frozen encoders.
All inherited and new parameters undergo embodied pretraining with independent video and action noise levels:
\begin{equation}
(\widehat{\mathbf{u}}_z,\widehat{\mathbf{u}}_a)=f_\theta(\mathbf{z}_{\tau_z},\mathbf{a}_{\tau_a},\tau_z,\tau_a\mid\xi),
\label{eq:wamflow}
\end{equation}
where $\mathbf{z}_{\tau_z}$ and $\mathbf{a}_{\tau_a}$ denote noisy video latents and actions, respectively, and $\xi=(\mathbf{o}_t,\ell,e)$.
Inspired by HBridge~\cite{hbridge}, layers 1--6 process the video--text and action streams separately.
In layers 7--26, action queries additionally attend to video and text tokens, giving
\begin{equation}
\widetilde{\mathbf{h}}_{cz}=\operatorname{Attn}(\mathbf{q}_{cz},\mathbf{k}_{cz},\mathbf{v}_{cz}),\;
\widetilde{\mathbf{h}}_a=\operatorname{Attn}(\mathbf{q}_a,\mathbf{k}_{cza},\mathbf{v}_{cza}),
\label{eq:actionattn}
\end{equation}
where $c,z,a$ denote text, video, and action, respectively, and combined subscripts indicate token concatenation.
Here, $\mathbf{q},\mathbf{k},\mathbf{v}$ are queries, keys, and values, and $\widetilde{\mathbf{h}}$ denotes the attention output.
Video and text tokens do not attend to actions.
Their representations nevertheless adapt to action supervision through gradients propagated along the key and value pathways.

\paragraph{Multi-view representation.}
Multi-view pretraining must accommodate heterogeneous cameras and resolutions while retaining video priors.
Tiling alters spatial layouts, scales poorly, and lacks explicit cross-view geometry.
We instead encode views independently and integrate their tokens through shared attention.
To accommodate resolution differences across datasets, view $k$ receives a target pixel area $s_k$:
\begin{equation}
\alpha_k=\sqrt{\frac{s_k}{h_kw_k}},\qquad
(\widetilde h_k,\widetilde w_k)=(\alpha_kh_k,\alpha_kw_k).
\label{eq:arearesize}
\end{equation}
Here, $h_k,w_k$ are the original frame dimensions, $\alpha_k$ is the resize factor, and $\widetilde h_k,\widetilde w_k$ are the resized dimensions.
This preserves aspect ratios before adjustment to multiples of 16 and controls token budgets independently of source resolution.
Our three-view pretraining configuration allocates target areas of 120,000 pixels to the main view and 30,000 pixels to each wrist view.

The encoder and token projection produce $\mathbf{h}_{z,k}^{(0)}=p_{\mathrm{vid}}(e_{\mathrm{vid}}(\widetilde{\mathbf{v}}^{(k)}))+\mathbf{b}_k$, where $\widetilde{\mathbf{v}}^{(k)}$ is the resized video from view $k$, $e_{\mathrm{vid}}$ is the video VAE encoder, $p_{\mathrm{vid}}$ projects latents into the transformer hidden space, and $\mathbf{b}_k$ is a zero-initialized camera embedding.
The resulting $\mathbf{h}_{z,k}^{(0)}$ denotes the input video tokens for that view.
Spatiotemporal RoPE operates independently within each view's local coordinates, while camera embeddings distinguish viewpoints.
Shared attention across concatenated tokens enables cross-view interaction and scales to varying camera configurations without redesigning tiled layouts.

\paragraph{Geometry conditioning.}
Frozen Depth Anything~3~\cite{da3} extracts features $\mathbf{g}^{(l)}$ from current observations, where $l$ indexes the corresponding WAM transformer block.
Trainable projections yield $\mathbf{k}_g^{(l)}=\mathbf{w}_{gk}^{(l)}\mathbf{g}^{(l)}$ and $\mathbf{v}_g^{(l)}=\mathbf{w}_{gv}^{(l)}\mathbf{g}^{(l)}$, augmenting action attention at $l\in\{10,12,\ldots,24\}$:
\begin{equation}
\widetilde{\mathbf{h}}_a^{(l)}
=\operatorname{Attn}\!\left(\mathbf{q}_a^{(l)},\mathbf{k}_{czag}^{(l)},\mathbf{v}_{czag}^{(l)}\right).
\label{eq:geometryattn}
\end{equation}
Here $g$ denotes geometry tokens.
This layer-wise fusion introduces spatial features at corresponding transformer depths, complementing the evolving video representations with observed geometry.

\subsection{Heterogeneous Embodied Pretraining}
\label{sec:embodied}
Our embodied pretraining corpus combines egocentric human interactions, Universal Manipulation Interface (UMI)~\cite{umi} trajectories, and demonstrations from diverse robot embodiments, including an in-house collection of robot manipulation data.
\paragraph{Shared action representation.}
Robot, human, and UMI trajectories share a 55D space for arm joints, EEF poses, grippers, torso, head, and chassis motion.
A validity mask $\mathbf{m}\in\{0,1\}^{h\times55}$ over the $h$-step horizon excludes padded entries from supervision and is concatenated with action inputs before projection.
Text conditioning combines the instruction, embodiment name, and action-layout description.
Arm joints are relative to the conditioning state.
EEF targets encode changes from the conditioning pose in that observation's reference camera frame, using 3D translation and continuous 6D rotation.
Human/UMI trajectories occupy the same EEF slots, with gripper supervision where available, avoiding joint retargeting.
Gripper/torso targets remain absolute; head/chassis targets are relative.
For action dimensions subject to normalization, each scalar target $a$ is transformed using its embodiment-specific 1st and 99th percentiles, $q_{0.01}$ and $q_{0.99}$,
\begin{equation}
\widetilde a=\operatorname{clip}\!\left(2\frac{a-q_{0.01}}{q_{0.99}-q_{0.01}}-1,\,-5,\,5\right).
\label{eq:actionquantile}
\end{equation}

\paragraph{Joint objective.}
We independently sample the video timestep $\tau_z$ and action timestep $\tau_a$ for each training sample.
Valid action entries follow Eq.~\eqref{eq:path} at $\tau_a$ with Gaussian noise $\boldsymbol{\eta}$ and target velocity $\mathbf{u}_a=\boldsymbol{\eta}-\mathbf{a}$.
The masked action loss is
\begin{equation}
\mathcal{L}_{\mathrm{act}}
=\mathbb{E}\!\left[
\frac{\sum_{k,j}m_{k,j}(\widehat u_{a,k,j}-u_{a,k,j})^2}
{\max(1,\sum_{k,j}m_{k,j})}
\right].
\label{eq:actionloss}
\end{equation}
Averaging valid entries prevents padding from reducing a sample's contribution.
We optimize $\mathcal{L}_{\mathrm{WAM}}=\mathcal{L}_{\mathrm{vid}}+\mathcal{L}_{\mathrm{act}}$, retaining per-clip video reduction and averaging both losses across samples.
Sources are sampled as $p(s)=p(g)d_s/\sum_{s'\in g}d_{s'}$, where $p(g)$ is the source-group probability and $d_s$ the recorded source duration.

\paragraph{Inference.}
For action-only inference, future-video latents remain fixed Gaussian noise at $\tau_z=1$, while observed-frame latents remain clean at $\tau_z=0$.
We denoise the action chunk from $\tau_a=1$ to $0$, without denoising or decoding future video frames.

%% file: sec/4_experiments.tex
\section{Experiments}
\label{sec:experiments}
We evaluate Gravis-Video on embodied video generation and Gravis-WAM on simulated manipulation, direct deployment, and multi-task adaptation, with ablations of embodied pretraining.

\subsection{Embodied Video Generation}
\label{sec:human}

\paragraph{Evaluation setup.}
We evaluate Gravis-Video on a 200-case embodied TI2V evaluation set covering robot manipulation from first-person or head-mounted views and egocentric human hand interactions.
Fourteen experts score anonymized, randomly presented generations from 1 to 5 on \emph{video quality (VQ)}---visual fidelity and temporal coherence; \emph{physical consistency (PC)}---plausibility of motion, contact, and object interactions; and \emph{embodiment consistency (EC)}---preservation of body appearance and structural integrity for robots and human hands.
Ratings are averaged within each case and then equally across cases.
We report mean scores with 95\% case-bootstrap confidence intervals.

\paragraph{Results.}
Gravis-Video scores 3.35 VQ, 3.21 PC, and 3.54 EC (Figure~\ref{fig:video_scores}).
Despite lower VQ than leading closed-source models, it achieves the highest mean EC and exceeds all evaluated open-weight baselines in PC and EC.
Thus, stronger general-purpose generation does not necessarily imply better embodied priors.
Figure~\ref{fig:video_generation} illustrates robot geometry and grasp contact alongside general text-to-video generation.

\input{tables/simulation}

\begin{figure}[!htb]
\begin{minipage}[t]{.46\textwidth}
\vspace{0pt}
\centering\small
\captionsetup*{type=table}
\caption{\textbf{Evaluation on EBench's 26 manipulation tasks.} Success rate (SR, \%) and stage-completion score (0--100). Gravis-WAM ranks first among the compared methods on both metrics.}
\label{tab:ebench}
\begin{tabularx}{\linewidth}{Ycc}
\toprule
Method & SR & Score\\
\midrule
PICA-WAM~\cite{picawam} & 61.4 & 74.9\\
OpenWAM-$\alpha$~\cite{openwamexp} & 55.3 & 70.0\\
InternW0-$\Delta$~\cite{internw0delta} & 49.2 & 66.0\\
GPT-6-Astra-ICL~\cite{astraebench} & 46.7 & 65.4\\
Qwen-RobotManip~\cite{qwenrobotmanip} & 45.6 & 60.8\\
$\pi_{0.5}$~\cite{pi05} & 41.4 & 54.4\\
GigaBrain-0.7~\cite{gigabrain07} & 33.3 & 46.0\\
\midrule
\rowcolor{paperpale}Gravis-WAM (Ours) & \textbf{62.5} & \textbf{76.1}\\
\bottomrule
\end{tabularx}
\end{minipage}\hfill
\begin{minipage}[t]{.50\textwidth}
\vspace{0pt}
\centering
\includegraphics[width=\linewidth]{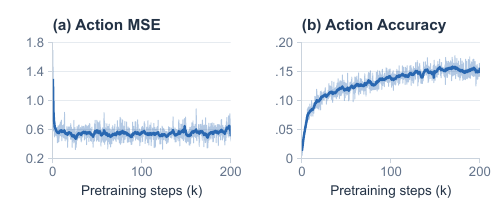}
\caption{\textbf{Action prediction on evaluation data during pretraining.} Normalized action MSE and accuracy over 200k steps. Accuracy counts valid action components with absolute error below $2/256$; both metrics exclude masked components.}
\label{fig:pretraining_curves}
\end{minipage}
\end{figure}

\subsection{Simulation Benchmark Evaluation}
\label{sec:sim}
We evaluate RoboTwin and EBench using benchmark metrics and published baselines; configurations are detailed in Appendix~\ref{app:bench}.

\paragraph{RoboTwin.}
We evaluate all 50 tasks~\cite{robotwin} with 50 trials per task in each evaluation condition.
Clean-only training uses 50 Clean demonstrations per task; Clean+Random uses 50 Clean and 500 Random demonstrations, sampled at 1:10.
Clean2Clean (C2C) and Clean2Random (C2R) evaluate Clean-only policies, with C2R testing randomization absent from downstream training.
Gravis-WAM achieves 87.08\% C2C and 63.40\% C2R, averaging 75.24\% and exceeding OpenWAM-$\alpha$ by 14.70 points on C2R (Table~\ref{tab:robotwin}).
Clean+Random training yields 94.60\% Clean and 93.84\% Random success; the 94.22\% mean leads the compared methods by 0.12 points (Table~\ref{tab:robotwinmixed}).

\paragraph{EBench.}
EBench~\cite{ebench} evaluates 26 tasks covering long-horizon, dexterous, and mobile manipulation.
We use four views---overhead, high, and two wrists---with target areas of 120,000 pixels for each global view and 30,000 for each wrist view; masked actions represent relative EEF poses, grippers, and planar base motion.
We reuse the pretrained main-view embedding for the overhead camera and retain both wrist embeddings, adding a zero-initialized learnable embedding for the high camera. Actions are denoised in 30 steps without future-video updates.
Gravis-WAM achieves 62.5\% success and 76.1 stage-completion score (0--100), exceeding the strongest compared baseline by 1.1 and 1.2 points, respectively (Table~\ref{tab:ebench}).
This demonstrates that the shared architecture accommodates additional views and a different control interface.

\subsection{Direct Deployment after Embodied Pretraining}
\label{sec:real}

\paragraph{Evaluation setup.}
We evaluate whether embodied pretraining enables real-world manipulation without task-specific adaptation.
Gravis-WAM is deployed directly from its pretrained checkpoint, whose heterogeneous training mixture includes approximately 1,000 hours of Astribot S1 demonstrations.
Each baseline retains its own pretrained foundation and is further trained on the same S1 corpus.

\input{tables/direct}

\begin{figure}[!htb]
\begin{minipage}[t]{\dimexpr.5\textwidth-.5\columnsep\relax}
\vspace{0pt}
\centering
\includegraphics[width=\linewidth]{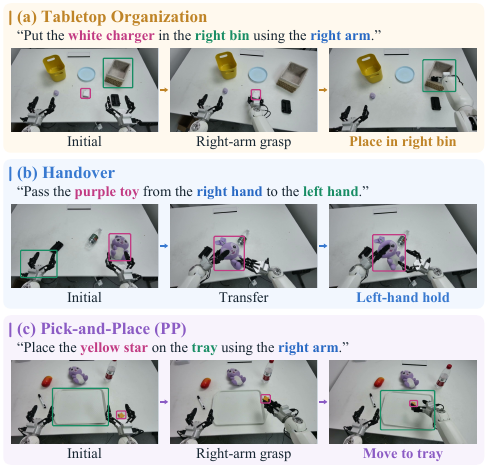}
\caption{\textbf{Instruction-guided direct deployment.} Without task-specific adaptation, Gravis-WAM follows arm, object, and destination instructions amid distractors: (a) tabletop organization, (b) handover, and (c) pick-and-place.}
\label{fig:pretraining_tasks}
\end{minipage}\hfill
\begin{minipage}[t]{\dimexpr.5\textwidth-.5\columnsep\relax}
\vspace{0pt}
\centering
\includegraphics[width=\linewidth]{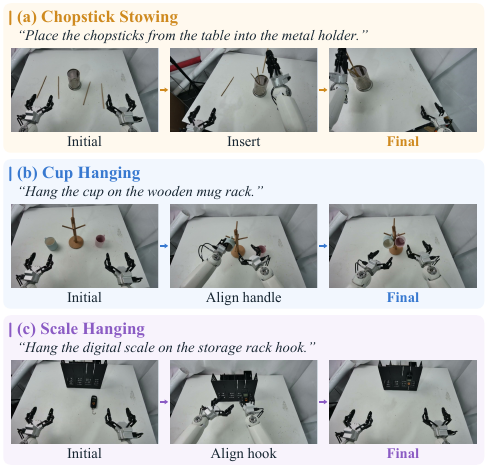}
\caption{\textbf{Joint multi-task adaptation.} One policy performs (a) chopstick stowing, (b) cup hanging, and (c) scale hanging after joint post-training on 310 demonstrations.}
\label{fig:posttraining_tasks}
\end{minipage}
\end{figure}

We evaluate Tabletop Organization, Handover, Pick-and-Place (PP), and Headphone Hanging with 36, 24, 20, and 20 trials, respectively.
The first three tasks are absent from the entire Gravis-WAM pretraining mixture; Headphone Hanging is represented in the S1 corpus.
The unseen tasks require identifying the instructed acting arm, target object, and destination among distractors (Figure~\ref{fig:pretraining_tasks}).
All methods share instructions, reset distributions, and success criteria; Overall is the unweighted task mean.

\paragraph{Results.}
Gravis-WAM achieves 80.42\% overall success, exceeding the strongest baseline by 5.69 percentage points (Table~\ref{tab:direct}).
Its clearest advantage is on Tabletop Organization, with 33 of 36 successes (91.67\%), demonstrating instruction-guided manipulation in multi-object scenes.
It also achieves 75\% on Handover, 80\% on PP, and a joint-best 75\% on Headphone Hanging, although it does not lead on Handover or PP.

\subsection{Multi-Task Post-Training}
\label{sec:post}

\paragraph{Evaluation setup.}
We jointly post-train one policy on Chopstick Stowing, Scale Hanging, and Cup Hanging using 56, 98, and 156 complete demonstrations (310 total).
All methods use the same data and 20 trials per task.
We report full-task success (SR) and partial success (PSR), averaging stage-completion scores with partial credit for completed stages.
Both metrics are averaged equally across tasks; details are in Appendix~\ref{app:real}.

\begin{figure}[!htb]
\centering
\includegraphics[width=\textwidth]{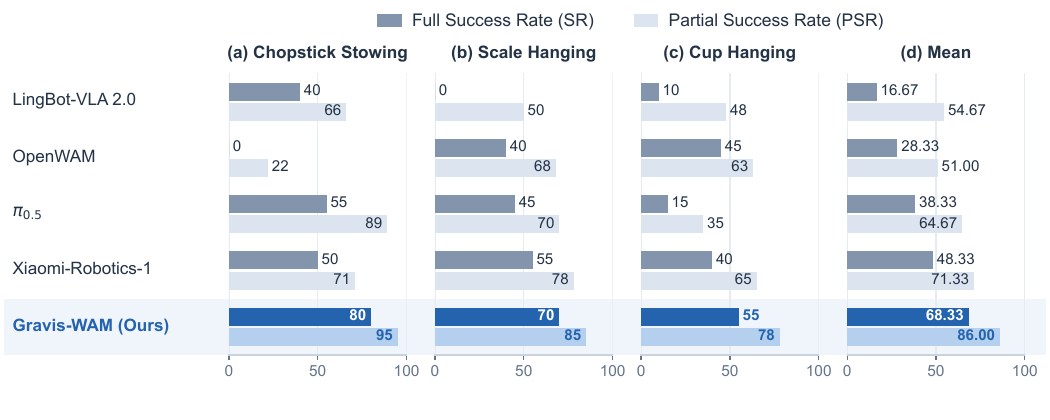}
\caption{\textbf{Three-task joint post-training.} Full success rate (SR; dark) and stage-based partial success rate (PSR; light), with unweighted task means. All methods use the same 310 demonstrations and 20 trials per task. Gravis-WAM leads both metrics across all three tasks, demonstrating effective real-world multi-task adaptation.}
\label{fig:posttraining_results}
\end{figure}

\paragraph{Results.}
Gravis-WAM leads SR and PSR on all three tasks, averaging 68.33\% and 86.00\%---20.0 and 14.7 points above the strongest baseline (Figure~\ref{fig:posttraining_results}).
Gains across distinct manipulation skills demonstrate effective transfer through joint multi-task adaptation (Figure~\ref{fig:posttraining_tasks}).

\begin{table}[!htbp]
\centering\small
\caption{\textbf{RoboTwin ablations and video-foundation comparison.} Success rates (\%) under Clean-only training, with embodied pretraining and DA3 progressively removed from Gravis-WAM. The full model performs best in both C2C and C2R. Wan2.2-based methods ($\dagger$) provide external comparisons under their respective architectures and training recipes.}
\label{tab:componentablation}
\begingroup
\setlength{\tabcolsep}{3pt}
\renewcommand{\arraystretch}{1.15}
\renewcommand{\tabularxcolumn}[1]{m{#1}}
\begin{tabularx}{\columnwidth}{@{}Y>{\raggedright\arraybackslash}m{.235\columnwidth}>{\centering\arraybackslash}m{.15\columnwidth}rr@{}}
\toprule
Method / Variant & Video backbone & Embodied PT & C2C & C2R\\
\midrule
\rowcolor{paperpale}\textbf{Gravis-WAM (Ours)} & \mbox{Gravis-Video} & $\checkmark$ & \textbf{87.08} & \textbf{63.40}\\
w/o embodied pretraining & \mbox{Gravis-Video} & $\times$ & 82.16 & 48.92\\
w/o embodied pretraining and DA3 & \mbox{Gravis-Video} & $\times$ & 81.28 & 50.36\\
\midrule
Fast-WAM~\cite{fastwam}$^\dagger$ & Wan2.2 TI2V-5B & $\times$ & 77.76 & 1.92\\
AHA-WAM~\cite{ahawam}$^\dagger$ & Wan2.2 TI2V-5B & $\times$ & 64.32 & 3.20\\
\bottomrule
\end{tabularx}
\endgroup
\end{table}

\subsection{Ablations and Video-Foundation Comparison}
\label{sec:embodiedablation}
\paragraph{Embodied pretraining.}
With the backbone and downstream setup fixed, embodied pretraining improves C2C by 4.92 points and C2R by 14.48 points (Table~\ref{tab:componentablation}), indicating stronger generalization beyond downstream training conditions.

\paragraph{Geometry conditioning.}
Without embodied pretraining or DA3, Gravis-WAM achieves 81.28\% C2C and 50.36\% C2R success, compared with 82.16\% and 48.92\% when retaining DA3.
These results show no consistent advantage from geometry conditioning in this setting and indicate that strong performance remains attainable without depth-derived features.

\paragraph{Video foundation.}
Direct adaptation of Gravis-Video with joint-space actions, without embodied pretraining or DA3, substantially outperforms both Wan2.2-based methods on C2R, supporting its effectiveness as a foundation for transferable embodied interaction priors.

%% file: tables/simulation.tex
\begin{table}[!htbp]
\centering\small
\setlength{\tabcolsep}{3pt}
\caption{\textbf{RoboTwin evaluation under Clean-only training.} Success rates (\%) on clean (C2C) and randomized (C2R) environments after training on clean demonstrations. Gravis-WAM achieves the highest average success across both conditions.}
\label{tab:robotwin}
\begin{tabularx}{\columnwidth}{Yccc}
\toprule
Method & C2C & C2R & Avg.\\
\midrule
StarVLA~\cite{starvla} & 46.52 & 3.16 & 24.84\\
GigaBrain-0.7~\cite{gigabrain07} & 66.80 & \textbf{67.92} & 67.36\\
Xiaomi Robotics-0~\cite{xiaomi0} & 62.86 & 18.22 & 40.54\\
$\pi_{0.5}$~\cite{pi05} & 70.66 & 46.00 & 58.33\\
Fast-WAM~\cite{fastwam} & 77.76 & 1.92 & 39.84\\
PrismWAM~\cite{prismwam} & 87.98 & 53.16 & 70.57\\
X-WAM~\cite{xwam} & 69.98 & 25.84 & 47.91\\
OpenWAM-$\alpha$~\cite{openwamexp} & \textbf{89.40} & 48.70 & 69.05\\
\midrule
\rowcolor{paperpale}Gravis-WAM (Ours) & 87.08 & 63.40 & \textbf{75.24}\\
\bottomrule
\end{tabularx}
\end{table}

\begin{table}[!htbp]
\centering\small
\setlength{\tabcolsep}{3pt}
\caption{\textbf{RoboTwin evaluation under joint Clean+Random training.} Success rates (\%) after training on both clean and randomized demonstrations. Gravis-WAM achieves 94.22\% average success, with strong performance in both environments.}
\label{tab:robotwinmixed}
\begin{tabularx}{\columnwidth}{Yccc}
\toprule
Method & Clean & Random & Avg.\\
\midrule
Motus~\cite{motus} & 88.66 & 87.02 & 87.84\\
Fast-WAM~\cite{fastwam} & 91.88 & 91.78 & 91.83\\
LingBot-VA~2.0~\cite{lingbotva2} & 93.80 & 93.40 & 93.60\\
Qwen-RobotManip~\cite{qwenrobotmanip} & 93.70 & 94.00 & 93.85\\
ABot-M0.5~\cite{abotm05} & 94.00 & \textbf{94.20} & 94.10\\
\midrule
\rowcolor{paperpale}Gravis-WAM (Ours) & \textbf{94.60} & 93.84 & \textbf{94.22}\\
\bottomrule
\end{tabularx}
\end{table}

%% file: tables/direct.tex
\begin{table}[!htbp]
\centering\small
\caption{\textbf{Direct deployment after embodied pretraining.} Task success (\%) and its unweighted mean (Overall). Gravis-WAM is deployed directly; baselines train on the same 1,000-hour S1 corpus included in Gravis's pretraining. PP denotes pick-and-place. Best results are bold.}
\label{tab:direct}
\begingroup
\setlength{\tabcolsep}{3pt}
\renewcommand{\arraystretch}{1.06}
\definecolor{directbarblue}{HTML}{9EBBD3}
\definecolor{directbaraccent}{HTML}{2464AF}
\definecolor{directbarpale}{HTML}{EAF2FE}
\newlength{\directbarunit}
\setlength{\directbarunit}{0.00215\textwidth}
\newcommand{\directsr}[2]{%  
\makebox[.265\textwidth][l]{%    
\textcolor{#2}{\raisebox{.5pt}{\rule{#1\directbarunit}{5.4pt}}}%    
\hspace{3pt}{\scriptsize #1}}}
\begin{tabularx}{\textwidth}{@{}Ycccc@{\hspace{9pt}}l@{}}
\toprule
\multicolumn{1}{@{}c}{Method} & \shortstack{Tabletop\\organization} & Handover & PP & \shortstack{Headphone\\hanging} & Overall success (\%)\\
\midrule
\piHalf~\cite{pi05} & 58.33 & \textbf{79.17} & 90 & 70 & \directsr{74.38}{directbarblue}\\
VPP 2.0~\cite{vpp2exp} & 72.22 & 66.67 & 85 & \textbf{75} & \directsr{74.72}{directbarblue}\\
LingBot-VLA 2.0~\cite{lingbotvla2exp} & 83.33 & 58.33 & 95 & 60 & \directsr{74.17}{directbarblue}\\
Motus~\cite{motus} & 75 & \textbf{79.17} & 85 & 55 & \directsr{73.54}{directbarblue}\\
Xiaomi-Robotics-1~\cite{xiaomiexp} & 75 & 37.5 & \textbf{100} & 35 & \directsr{61.88}{directbarblue}\\
OpenWAM~\cite{openwamexp} & 66.67 & 70.83 & 65 & 25 & \directsr{56.88}{directbarblue}\\
GR00T N1.7~\cite{gr00tn1_2025} & 19.44 & 45.83 & 75 & 5 & \directsr{36.32}{directbarblue}\\
\midrule
\rowcolor{directbarpale}Gravis-WAM (Ours) & \textbf{91.67} & 75 & 80 & \textbf{75} & \textbf{\directsr{80.42}{directbaraccent}}\\
\bottomrule
\end{tabularx}
\endgroup
\end{table}

%% file: sec/6_conclusion.tex
% Allow the conclusion to follow the preceding text naturally.
\section{Conclusion and Limitations}
\label{sec:conclusion}
\label{sec:analysis}
\label{sec:limitations}
We presented Gravis, a two-stage embodied-native pretraining framework.
Gravis-Video learns visual interaction priors from curated real-world videos, while Gravis-WAM aligns heterogeneous embodied data through scalable native multi-view modeling, depth-derived spatial context, unified actions, and embodiment conditioning.
Video evaluation highlights that perceptual quality alone does not guarantee embodiment consistency.
Experiments on RoboTwin, EBench, and real robots demonstrate strong simulated manipulation, direct deployment on unseen semantically demanding tasks, and effective real-world multi-task adaptation.
Ablations further demonstrate that embodied pretraining improves generalization beyond downstream training conditions.
Together, these findings support learning embodied visual priors and aligning heterogeneous experience as complementary foundations for generalizable WAMs.
        
Computational constraints leave controlled video-data ablations and model and data scaling for future work.

%% file: sec/supplementary.tex
\appendix
\titleformat*{\paragraph}{\normalfont\normalsize\bfseries}
\titlespacing*{\paragraph}{0pt}{8pt plus 2pt minus 1pt}{.65em}
% Supplement-only float placement: keep large tables and figures near their discussion.
\setcounter{topnumber}{4}
\setcounter{totalnumber}{6}
\setcounter{bottomnumber}{3}
\setcounter{dbltopnumber}{4}
\renewcommand{\topfraction}{.95}
\renewcommand{\dbltopfraction}{.95}
\renewcommand{\textfraction}{.05}
\renewcommand{\bottomfraction}{.9}
\renewcommand{\floatpagefraction}{.82}
\renewcommand{\dblfloatpagefraction}{.82}
\makeatletter
\setlength{\@fptop}{0pt}
\setlength{\@dblfptop}{0pt}
\setlength{\@fpsep}{12pt plus 2pt minus 2pt}
\setlength{\@dblfpsep}{12pt plus 2pt minus 2pt}
\makeatother

\section{Data Curation and Preprocessing}
\label{app:data}

\subsection{Internet Video Filtering}
We filter the Internet corpus along two axes: content realism and temporal continuity.
Content filtering removes animation, simulation, rendered 3D footage, and videos with visual effects that distort physical appearance or motion.
Temporal filtering excludes shot transitions, time-lapse footage, and slow-motion sequences that disrupt continuity or alter the natural motion rate.
These filtering criteria apply to the Internet-video corpus.

\subsection{Video Preprocessing}
Video frames are resized with a target long side of 512 pixels for PT and 896 pixels for CT.
The short side follows the original aspect ratio and is rounded down to a multiple of 16 through resizing and center cropping.
For example, a 16:9 frame becomes $512\times288$ in PT and $896\times496$ in CT.
Both stages cap each sample at 100,000 latent tokens to exclude excessively long clips and bound per-sample computational cost.

\subsection{Embodied Pretraining Data}
Our embodied pretraining corpus comprises approximately 370,000 hours of robot demonstrations, egocentric human interactions, and UMI trajectories after validation splitting.
Robot demonstrations span multiple embodiments and include internally collected manipulation data.
Egocentric human trajectories provide hand-derived end-effector supervision, while UMI trajectories provide recorded end-effector poses and available gripper measurements.
Table~\ref{tab:data} summarizes the duration of each category.

\begin{table}[!htbp]
\centering\small
\caption{\textbf{Embodied pretraining data by category.} Approximate durations count each episode's primary view once, without multiplying by camera count or sampling weight.}
\label{tab:data}
\begin{tabularx}{\columnwidth}{Y r}
\toprule
Data category & \shortstack{Approximate duration\\(k hours)}\\
\midrule
Egocentric human interactions & 267\\
UMI trajectories & 82\\
Robot demonstrations & 22\\
\midrule
\textbf{Total} & $\boldsymbol{\approx370}$\\
\bottomrule
\end{tabularx}
\end{table}

Reported durations reflect the collected recordings before filtering for valid action supervision.

\subsection{Embodied Sequence Preparation and Sampling}
Each training window contains one observed conditioning frame and 12 future video frames, paired with 24 action steps sampled at two steps per inter-frame interval.
Video frames are sampled at strides of six for robot trajectories and four for egocentric human and UMI trajectories, measured on the source frame timeline.
Absolute sampling frequencies therefore depend on the source frame rate.

We sample training data in two steps: first choosing a data category, then selecting a source within that category.
Sources with more recorded hours are sampled more frequently, with probabilities proportional to their recording durations.

\section{Gravis-Video Architecture and Pretraining}
\label{app:training}

\subsection{Architecture and Conditioning}
Gravis-Video uses modality-specific input projections to map video latents, audio latents, and text features into a shared hidden dimension.
Each transformer block maintains separate attention projections and gated feed-forward networks for the three streams.
Table~\ref{tab:videoarchitecture} summarizes the architecture.

\begin{table}[!htbp]
\centering\small
\caption{\textbf{Gravis-Video architecture.} Parameter counts exclude external modality encoders.}
\label{tab:videoarchitecture}
\begin{tabularx}{\columnwidth}{Y r}
\toprule
Configuration & Value\\
\midrule
Transformer blocks & 26\\
Hidden dimension & 2,560\\
Attention heads & 20\\
Dimension per head & 128\\
Feed-forward intermediate dimension & 6,912\\
Video latent channels & 48\\
Audio latent channels & 64\\
Text feature dimension & 3,584\\
Total parameters & 8.20B\\
Video and text pathways with shared time conditioning & 5.45B\\
\bottomrule
\end{tabularx}
\end{table}

Video attention uses 3D RoPE, allocating 32, 48, and 48 dimensions per head to the temporal, height, and width axes, respectively.
Audio attention applies 1D RoPE to 32 dimensions per head.
Queries and keys are RMS-normalized before attention.

The timestep $\tau$ is multiplied by 1,000 and encoded by a 256-dimensional sinusoidal embedding, followed by a two-layer MLP with SiLU activation.
Each block maps this embedding to modality-specific feature scales and residual gates.
Video modulation is shared across spatial tokens within each latent frame, whereas audio modulation is applied per token.

\subsection{Pretraining Configuration}
Gravis-Video is pretrained from scratch for 47,500 updates, followed by 4,500 continued-training updates at higher resolution.
CT starts from the PT EMA checkpoint, and its resulting EMA checkpoint initializes Gravis-WAM.
Both stages use the optimization settings in Table~\ref{tab:videopretraining}.
After 1,000 warmup updates, the learning rate remains constant.
Data filtering, resolution processing, and clip-length limits are described in Appendix~\ref{app:data}.

\begin{table}[!htbp]
\centering\small
\caption{\textbf{Training configurations for Gravis-Video.} PT and CT denote initial pretraining and higher-resolution continued training, respectively.}
\label{tab:videopretraining}
\begin{tabularx}{\columnwidth}{Y r r}
\toprule
Configuration & PT & CT\\
\midrule
Training steps & 47,500 & 4,500\\
Target frame long side & 512 & 896\\
Global batch size & 2,048 & 2,048\\
Learning rate & $2\times10^{-4}$ & $2\times10^{-4}$\\
Warmup updates & 1,000 & 1,000\\
Optimizer $(\beta_1,\beta_2)$ & $(0.9,0.95)$ & $(0.9,0.95)$\\
Weight decay & 0.01 & 0.01\\
Gradient clipping threshold & 1.0 & 1.0\\
Text-condition dropout & 0.1 & 0.1\\
Maximum EMA decay & 0.9999 & 0.9999\\
T2V:TI2V sampling ratio & 8:2 & 8:2\\
\bottomrule
\end{tabularx}
\end{table}

Noise levels follow a resolution-dependent logit-normal distribution.
For clip $i$ with latent spatial dimensions $h_i\times w_i$, we sample
\begin{equation}
\begin{aligned}
\tau_i&=\operatorname{sigmoid}(\epsilon_i+\log r_i),
&\epsilon_i&\sim\mathcal{N}(0,1),\\
r_i&=\min\!\left(6,\;3\sqrt{\frac{h_iw_i}{576}}\right).
\end{aligned}
\label{eq:video_resolution_timeshift}
\end{equation}
Here, $r_i$ is the time-shift factor, and 576 is the reference latent spatial area.
This adjustment shifts higher-resolution samples toward larger noise levels.
The sampled $\tau_i$ is shared by the predicted video and accompanying audio, while conditioning frames remain clean.

Variable-length sequences are packed for efficient computation, with attention restricted to tokens from the same example.
Losses are reduced separately for each clip before batch aggregation, retaining the per-example weighting defined in the main text.

\section{Gravis-WAM Architecture and Pretraining}
\label{app:wam_training}

\subsection{Model Construction and Spatial Conditioning}
\paragraph{Parameter transfer.}
Gravis-WAM retains the 26-block architecture described in Appendix~\ref{app:training}.
Initialization loads the complete video and text pathways, including their attention projections, feed-forward networks, and shared timestep embedding.
Audio-specific parameters are excluded; the action stream, geometry projections, and camera embeddings are initialized separately.
The inherited backbone remains trainable during embodied pretraining.

\paragraph{Multi-view implementation.}
Each available camera stream is independently resized and VAE-encoded.
Three-view samples allocate 120,000 pixels to the main camera and 30,000 pixels to each wrist camera; single-view samples allocate 180,000 pixels to their sole view.
Resizing uses bicubic interpolation and adjusts spatial dimensions to multiples of 16.

Packed sequences retain separate boundaries for individual views and complete training examples.
View boundaries define independent local RoPE coordinates, while example boundaries permit attention across views from the same example and prevent attention across different examples.
A zero-initialized camera embedding is added after the video input projection and shared across all spatial and temporal tokens from that camera.
Samples contain only their available views, without inserting blank videos for absent cameras.

\paragraph{Geometry feature extraction and injection.}
The frozen DA3 backbone jointly processes the observed images from all available cameras.
During pretraining, each image is bilinearly resized to $280\times504$ pixels and normalized using ImageNet statistics.
With a patch size of 14, each view contributes 720 patch features of dimension 1,024.
Only backbone features are extracted; the depth and camera prediction heads are not executed.

We extract features from DA3 global-attention layers $l\in\{10,12,\ldots,24\}$ and inject them into the corresponding WAM layers.
These features have already incorporated cross-view information.
Each injection module projects the 1,024-dimensional features to the 2,560-dimensional WAM hidden space, followed by a projection to keys and values.
Geometry keys undergo per-head RMS normalization before concatenation with the existing action-attention keys.
The geometry branch does not add WAM camera embeddings or apply an additional RoPE transformation.

\subsection{Shared Action Representation}
\paragraph{Action layout and masking.}
Table~\ref{tab:canonical55_details} specifies the shared action layout using zero-based indices.
Robot trajectories can supervise joint and EEF fields jointly when both are available; human and UMI trajectories populate the corresponding EEF fields and available gripper fields.

\begin{table}[!htbp]
\centering\small
\caption{\textbf{Shared 55-dimensional action representation.} Indices are zero-based. Relative targets are referenced to the conditioning-time state, and unavailable dimensions are masked.}
\label{tab:canonical55_details}
\begin{tabularx}{\textwidth}{lYcl}
\toprule
Indices & Field & Dimensions & Representation\\
\midrule
0--6 & Left-arm joints & 7 & Relative\\
7--15 & Left EEF pose & 9 & Relative, camera-referenced\\
16 & Left gripper & 1 & Absolute\\
17--23 & Right-arm joints & 7 & Relative\\
24--32 & Right EEF pose & 9 & Relative, camera-referenced\\
33 & Right gripper & 1 & Absolute\\
34--37 & Torso joints & 4 & Absolute\\
38--40 & Reserved torso slots & 3 & Padding\\
41--49 & Torso EEF pose & 9 & Relative, camera-referenced\\
50--51 & Head joints & 2 & Relative\\
52--54 & Planar chassis motion & 3 & Relative\\
\bottomrule
\end{tabularx}
\end{table}

Before flow interpolation, invalid action entries and their sampled Gaussian noise are set to zero.
At each prediction step, the noisy 55-dimensional action vector is concatenated with its 55-dimensional binary mask, producing a 110-dimensional input to the action projection.
The output head predicts 55-dimensional velocities; the mask is supplied as conditioning rather than predicted.

\paragraph{Relative joint targets.}
Arm and head joint targets are computed relative to the conditioning-time configuration:
\begin{equation}
\Delta\mathbf q_j=\mathbf q_j-\mathbf q_0,
\label{eq:relative_joint_details}
\end{equation}
where $\mathbf q_0$ denotes the conditioning-time joint positions and $\mathbf q_j$ denotes the future positions at prediction step $j$.
All future targets share the same reference rather than representing differences between consecutive steps.
Torso joint targets remain absolute.
These transformations precede statistical normalization.

\paragraph{Camera-referenced relative EEF targets.}
Let $(\mathbf p_0,\mathbf r_0)$ denote the EEF pose at the conditioning time, $(\mathbf p_j,\mathbf r_j)$ a future target pose, and $\mathbf r_c$ the rotation mapping the reference camera frame into their common coordinate frame.
The targets are constructed as
\begin{equation}
\begin{aligned}
\Delta\mathbf p_j&=\mathbf r_c^\top(\mathbf p_j-\mathbf p_0),\\
\Delta\mathbf r_j&=\mathbf r_c^\top\mathbf r_j\mathbf r_0^\top\mathbf r_c.
\end{aligned}
\label{eq:relative_eef_details}
\end{equation}
Each EEF target concatenates the three-dimensional translation with the first two columns of $\Delta\mathbf r_j$, yielding a continuous six-dimensional rotation representation.
The reference pose and camera transform are used to construct training targets; they are not supplied as additional state inputs to the transformer.
Human hand trajectories use virtual EEF poses derived from tracked hand landmarks.

Chassis translation is expressed relative to the conditioning-time chassis heading, and the yaw difference is wrapped to $[-\pi,\pi]$.

\paragraph{Normalization details.}
We use the quantile normalization defined in Eq.~\eqref{eq:actionquantile}, with embodiment-specific statistics.
Relative fields use prediction-step-specific statistics, whereas absolute fields share statistics across the horizon.
Dimensions excluded from normalization or with an interquantile range below $2\times10^{-3}$ remain unchanged.
The action-validity mask independently determines which entries contribute to supervision.

\begin{figure}[!htbp]
\centering
\includegraphics[width=\textwidth]{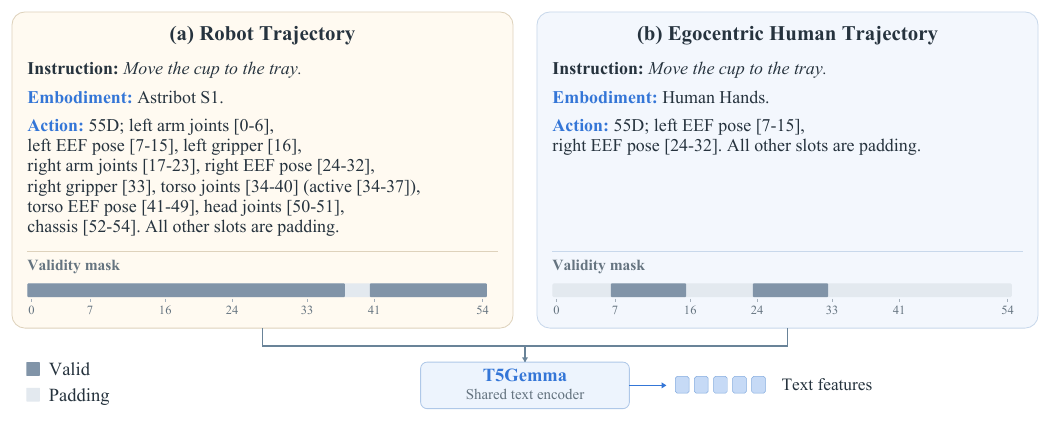}
\caption{\textbf{Embodiment-aware text conditioning.} Robot and human examples combine task instructions with embodiment identity and active fields in the shared action representation. A separate binary mask identifies valid action dimensions.}
\label{fig:appendix_text_conditioning}
\end{figure}

\paragraph{Embodiment-aware text construction.}
The text template contains the instruction, embodiment name, and active action fields with their canonical index ranges.
A dimension appears in the description if it is valid at any prediction step.
Contiguous active dimensions are grouped into ranges, and partially available fields explicitly list their active subranges.
Figure~\ref{fig:appendix_text_conditioning} illustrates the template for robot and egocentric human trajectories.

\subsection{Embodied Pretraining and Validation Details}
\paragraph{Training configuration.}
Table~\ref{tab:training} summarizes the embodied pretraining configuration.
The video VAE and geometry encoder remain frozen, while the inherited transformer and newly introduced WAM parameters are jointly optimized.
After warmup, the learning rate follows a cosine schedule with a minimum value equal to one-sixth of the peak learning rate.
Training uses the joint video--action objective defined in the main text.

\begin{table}[!htbp]
\centering\small
\caption{\textbf{Embodied pretraining configuration for Gravis-WAM.}}
\label{tab:training}
\begin{tabularx}{\columnwidth}{Y r}
\toprule
Configuration & Value\\
\midrule
Training updates & 200,000\\
Global batch size & 2,048\\
Peak learning rate & $8\times10^{-5}$\\
Warmup updates & 5,000\\
Learning-rate schedule & Cosine\\
Minimum learning-rate factor & $1/6$\\
Optimizer $(\beta_1,\beta_2)$ & $(0.9,0.95)$\\
Weight decay & 0.01\\
Text-condition dropout & 0.1\\
Maximum EMA decay & 0.9999\\
Video loss weight & 1.0\\
Action loss weight & 1.0\\
\bottomrule
\end{tabularx}
\end{table}

\paragraph{Independent noise sampling.}
Video and action noise levels are independently sampled for each example:
\begin{equation}
\begin{aligned}
\tau_z&=\operatorname{sigmoid}(\epsilon_z+\log 6),\\
\tau_a&=\operatorname{sigmoid}(\epsilon_a),\qquad
\epsilon_z,\epsilon_a\overset{\mathrm{i.i.d.}}{\sim}\mathcal N(0,1).
\end{aligned}
\label{eq:wam_noise_sampling_details}
\end{equation}
All predicted views within an example share $\tau_z$, and all valid action entries share $\tau_a$.
The two streams reuse the pretrained timestep embedding with their respective noise levels.

\paragraph{Observation augmentation.}
Training applies brightness, contrast, and saturation jitter with strength 0.15.
Pixel-space conditioning noise is enabled with probability 0.5 and a configured strength range of $[0,0.05]$.
Independently, latent-space conditioning noise is applied with probability 0.5, using a noise level sampled uniformly from $[0.3,0.7]$.
When latent noise is applied, the observed-frame timestep embedding records its actual noise level.
Observed frames remain excluded from the video prediction loss.
DA3 receives the observed RGB images before latent-space corruption.

\paragraph{Validation metrics.}
Figure~\ref{fig:pretraining_curves} reports open-loop action prediction in the \textbf{normalized action space}.
After diffusion sampling, predicted action chunks are compared with demonstration targets using the same embodiment-specific normalization.
Metrics are computed before inverse normalization or conversion to executable robot coordinates.
Let $\widehat{\widetilde a}_{ijd}$ and $\widetilde a_{ijd}$ denote the predicted and ground-truth normalized action components for example $i$, prediction step $j$, and dimension $d$, respectively.
With validity mask $m_{ijd}$, the reported metrics are
\begin{equation}
\mathrm{MSE}
=
\frac{\sum_{i,j,d}m_{ijd}
\left(\widehat{\widetilde a}_{ijd}-\widetilde a_{ijd}\right)^2}
{\sum_{i,j,d}m_{ijd}},
\label{eq:normalized_validation_mse}
\end{equation}
\begin{equation}
\mathrm{Accuracy}
=
\frac{\sum_{i,j,d}m_{ijd}\,
\mathbf{1}\!\left[
\left|\widehat{\widetilde a}_{ijd}-\widetilde a_{ijd}\right|
<\frac{2}{256}
\right]}
{\sum_{i,j,d}m_{ijd}}.
\label{eq:normalized_validation_accuracy}
\end{equation}
Error sums and valid-entry counts are accumulated across evaluation examples and devices before division.
Accuracy measures the fraction of valid normalized action components with absolute error below $2/256$; it does not represent task success.
These metrics assess sampled action predictions, whereas the training losses measure flow-velocity prediction errors.

% Training-loss curves will be inserted here once the original metric series is available.
% Do not substitute validation action-prediction metrics for training losses.

\section{Simulation Benchmark Details and Per-Task Results}
\label{app:bench}

\subsection{RoboTwin}
\input{tables/robotwin_per_task_appendix}

\paragraph{Training and evaluation.}
Clean-only training uses 50 demonstrations per task; Clean+Random combines 50 Clean and 500 Random demonstrations with a sampling ratio of 1:10.
Both policies start from the 200k-step embodied-pretrained checkpoint. Reported results use the 20k-step and 180k-step downstream EMA checkpoints, respectively.
Each setting evaluates 50 trials on each of the 50 tasks, using benchmark-defined success criteria.
Both runs use global batch 128, optimizer betas $(0.9,0.95)$, and weight decay 0.01.

\paragraph{Observation and action interfaces.}
Each window contains one conditioning frame and 12 future frames, with two action steps per future frame.
The Clean-only policy predicts camera-frame relative EEF poses and absolute gripper commands in the masked 55-dimensional representation, with 20 active components.
Predictions are inverse-normalized using saved training statistics and converted to world-frame poses for RoboTwin's end-effector control interface.
Clean+Random instead uses 14 unnormalized components: six absolute joint positions and one continuous gripper command per arm.
Its action projections are reinitialized, and mask concatenation is disabled.

\paragraph{Inference.}
\label{app:inference}
We jointly denoise future video latents and 24-step action chunks using 30-step Euler integration, with time-shift factors of 6 for video and 1 for actions.
Observed-frame latents remain fixed; future video latents are not decoded into pixels.
Table~\ref{tab:robotwin_per_task} reports per-task results under both training protocols.

\subsection{EBench}
\paragraph{Four-view adaptation.}
We expand the pretrained view-embedding table from three to four entries.
The overhead camera inherits the main-view embedding, both wrist embeddings are retained, and the high camera receives a new zero-initialized embedding; all four are optimized during training.
The global and wrist views receive area budgets of 120,000 and 30,000 pixels, respectively.
All views are required, and geometry inputs are aspect-ratio padded to $448\times336$ pixels.

\paragraph{Actions and optimization.}
The masked 55-dimensional representation supervises 22 components: two relative EEF poses, two absolute gripper values, and planar base displacements.
Gripper targets average the two finger measurements; base yaw is excluded.
Relative EEF targets remain in the dataset coordinate frame without an additional reference-camera transformation.
Training uses sub-instructions and windows of one conditioning frame and 12 future frames, with three action steps per future frame.
Starting from the 200k-step embodied-pretrained checkpoint, we reinitialize the action input-projection weight and output projection and train for 200k updates with global batch 128.
The learning rate is $2\times10^{-5}$ after 100 warmup steps, with no decay; optimizer betas are $(0.9,0.95)$ and weight decay is 0.01.
View dropping and image-color augmentation are disabled; latent conditioning noise is applied with probability 0.5 at levels sampled uniformly from $[0.3,0.7]$.

\input{tables/ebench_per_task_appendix}

\paragraph{Inference.}
We jointly denoise future video latents and 36-step action chunks using 30 denoising steps, with an action time-shift factor of 1.
Observed-frame latents remain fixed; future video latents are not decoded into pixels.
\paragraph{Per-task results.}
Tables~\ref{tab:ebench_per_task} and~\ref{tab:ebench_per_task_score} report success rates and stage-completion scores for the same baseline set as Table~\ref{tab:ebench}.
Gravis-WAM is evaluated using its 100k-step downstream EMA checkpoint on 510 episodes across 26 tasks: Make Sandwich and Microwave have 15 trials each, and all other tasks have 20.
All 510 episode records are complete and unique.
We report task-equal means, yielding 62.50\% SR and a score of 76.15; pooling episodes instead yields 62.55\% (319/510) and 76.02.
Baseline task values are taken from the public leaderboard.
Means in these tables are recomputed from the exported per-task values; the leaderboard's separately reported aggregate scores can differ slightly.
Gravis-WAM reaches 100\% success on Apple To Fruit Bowl, Detergent, and Salt To Spice Rack, while Bottle, Pen, and Collect Coffee Beans remain challenging.

\Needspace{7\baselineskip}
\section{Real-World Tasks and Evaluation Procedures}
\label{app:real}
\subsection{Direct-Deployment Tasks}
\label{app:direct_tasks}

\paragraph{Tabletop Organization.}
The scene contains several small objects, including a white charger, a purple toy, and a black phone holder, together with multiple possible destinations: a yellow storage bin, a woven basket, and a blue plate.
The robot must identify the instructed object, grasp it with the specified arm, and place it in the designated container or on the designated surface.
For example, an instruction may request placing the white charger in the right-hand bin using the right arm.
The presence of multiple objects and destinations requires resolving both object identity and spatial references.

\paragraph{Handover.}
Objects such as a water bottle, a purple plush toy, and a phone holder are placed on the table.
The robot must select the instructed object, grasp it with the designated source arm, and transfer it to the receiving hand.
For example, the right hand picks up the purple toy and passes it to the left hand.
Completion requires the receiving hand to retain the object after the source hand releases it, combining instruction understanding with coordinated bimanual manipulation.

\paragraph{Pick-and-Place.}
A tray is surrounded by candidate objects, including a marker, a water bottle, a purple plush toy, a yellow star-shaped block, and a fruit-shaped object.
The robot must use the instructed arm to select the target and place it on the tray.
For example, the left arm retrieves the marker while ignoring the other objects.
This task tests target selection among distractors with diverse appearances, shapes, and grasp requirements.

\paragraph{Headphone Hanging.}
The scene contains a pair of headphones and a designated support.
The robot must pick up the headphones, orient them appropriately, and hang them on the support.
Successful completion requires the headphones to remain supported after the gripper releases them, demanding coordinated grasping, orientation adjustment, and placement.

\subsection{Joint Multi-Task Post-Training}
\label{app:post_tasks}
\paragraph{Tasks and scoring.}
Table~\ref{tab:post_detailed} defines the completion stages and partial-credit scores for Chopstick Stowing, Scale Hanging, and Cup Hanging.
Each task is evaluated over 20 trials.
SR is the percentage of fully successful trials; PSR is the mean completion score across all trials, expressed as a percentage.
Overall metrics are averaged equally across tasks.

\paragraph{Initialization.}
Each baseline first continues training its own pretrained checkpoint on the approximately 1,000-hour Astribot S1 corpus described in Section~\ref{sec:real}, and then uses this checkpoint for three-task joint post-training.
Gravis-WAM starts from its embodied-pretrained checkpoint, whose training mixture includes the same S1 corpus.
All methods therefore have prior training exposure to the target embodiment; Gravis-WAM's stronger results under this protocol support effective transfer from embodied pretraining to multi-task adaptation.

\paragraph{Joint adaptation.}
We jointly post-train Gravis-WAM on 56 Chopstick Stowing, 98 Scale Hanging, and 156 Cup Hanging demonstrations, sampling tasks uniformly.
We use a global batch size of 128 and evaluate the EMA checkpoint at 5,000 updates.

Table~\ref{tab:post_detailed} reports stage-wise completion counts and the corresponding SR and PSR for each model.

\begin{table}[!htbp]
\centering\footnotesize
\caption{\textbf{Completion counts and partial-credit scores after joint multi-task post-training.} Each model is evaluated over 20 trials per task. Trials are counted once at their final completion level, with scores shown in parentheses. SR measures full completion, while PSR averages completion scores.}
\label{tab:post_detailed}
\begingroup
\setlength{\tabcolsep}{5pt}
\renewcommand{\arraystretch}{1.0}
\begin{tabularx}{\textwidth}{@{}Yccccc@{}}
\toprule
Final completion category (score) & \shortstack{OpenWAM} & \shortstack{\piHalf} & \shortstack{LingBot-VLA 2.0} & \shortstack{Xiaomi-Robotics-1} & \shortstack{\textbf{Gravis-WAM}\\\textbf{(Ours)}}\\
\midrule
\rowcolor{paperpale}\multicolumn{6}{c}{\textit{Chopstick Stowing}}\\
No chopsticks stowed (0) & 8 & 0 & 1 & 1 & 0 \\
One chopstick stowed (0.25) & 6 & 0 & 4 & 4 & 0 \\
Two chopsticks stowed (0.50) & 6 & 0 & 4 & 2 & 0 \\
Three chopsticks stowed (0.75) & 0 & 9 & 3 & 3 & 4 \\
All four chopsticks stowed (1.00) & 0 & 11 & 8 & 10 & 16\\
\cmidrule(lr){1-6}
SR (\%) & 0 & 55 & 40 & 50 & \textbf{80}\\
PSR (\%) & 22.5 & 88.75 & 66.25 & 71.25 & 95 \\
\midrule
\rowcolor{paperpale}\multicolumn{6}{c}{\textit{Scale Hanging}}\\
Scale not picked up (0) & 1 & 1 & 0 & 0 & 0\\
Scale picked up but not hung (0.50) & 11 & 10 & 20 & 9 & 6\\
Scale successfully hung (1.00) & 8 & 9 & 0 & 11 & 14\\
\cmidrule(lr){1-6}
SR (\%) & 40 & 45 & 0 & 55 & \textbf{70}\\
PSR (\%) & 67.5 & 70 & 50 & 77.5 & 85\\
\midrule
\rowcolor{paperpale}\multicolumn{6}{c}{\textit{Cup Hanging}}\\
No cups hung (0) & 4 & 9 & 3 & 2 & 0 \\
One cup hung (0.50) & 7 & 8 & 15 & 10 & 9 \\
Both cups hung (1.00) & 9 & 3 & 2 & 8 & 11\\
\cmidrule(lr){1-6}
SR (\%) & 45 & 15 & 10 & 40 & \textbf{55}\\
PSR (\%) & 62.5 & 35 & 47.5 & 65 & 77.5 \\
\bottomrule
\end{tabularx}
\endgroup
\end{table}

\subsection{Deployment Implementation}
\label{app:real_deployment}
Gravis-WAM receives observations from the head camera and two wrist cameras, using the same view-specific preprocessing as training.
Inference uses BF16 mixed precision and 30-step Euler integration, with time-shift factors of 6 for video and 1 for actions.
Each prediction produces a 24-step action chunk through joint video--action denoising, while observed-frame latents remain fixed.
Predicted relative EEF poses are inverse-normalized and converted into executable robot poses using the reference robot state and camera calibration.
New observations are acquired for subsequent predictions, enabling closed-loop execution.

\FloatBarrier
\section{Embodied Video Evaluation}
\label{app:human}
\paragraph{Annotation platform.}
The platform displays a real reference video alongside generated videos under anonymized model slots.
Each evaluator sees the models in a random order for each case.
Evaluators independently assign integer scores from 1 to 5 for video quality, physical consistency, and embodiment consistency.
For human-hand videos, embodiment consistency concerns hand anatomy and structural stability; for robot videos, it concerns robot morphology and articulation.

\noindent\begin{minipage}[t]{\textwidth}
\captionsetup{type=table}
\centering\small
\caption{\textbf{Human evaluation criteria for embodied video generation.} Experts independently rate video quality, physical consistency, and embodiment consistency on a 1--5 scale, with higher scores indicating better performance.}
\label{tab:rubric}
\begin{tabularx}{\textwidth}{lYY}
\toprule
Criterion & Assessed properties & Typical failure modes\\
\midrule
Video quality (VQ) & Visual fidelity and temporal coherence. & Blur, flicker, and rendering artifacts.\\
Physical consistency (PC) & Plausible motion, contact, support, and object persistence. & Interpenetration, unsupported motion, and disappearing objects.\\
Embodiment consistency (EC) & Stable appearance, morphology, and articulation of the acting body. & Changing limb counts or proportions, disconnected parts, and inconsistent joint structure.\\
\bottomrule
\end{tabularx}
\end{minipage}\par\medskip

\paragraph{Score aggregation.}
Fourteen experts evaluate the 200 cases.
Only submitted ratings containing all three scores are included.
For each model, ratings are first averaged within each case, then equally across cases.
The 95\% confidence intervals in Figure~\ref{fig:video_scores} are computed from 50,000 bootstrap resamples of the cases, retaining all associated ratings within each resampled case.

\section{Additional Qualitative Results}
\label{app:qual}
\subsection{General Text-to-Video Generation}
Figure~\ref{fig:appendix_general_t2v} provides two additional text-only generations: a fisherman repairing a net and a baker kneading dough.
These examples illustrate general scene synthesis and human--object interaction beyond robot manipulation.
Both examples use the CT checkpoint.

\noindent\begin{minipage}[t]{\textwidth}
\captionsetup{type=figure}
\centering
\includegraphics[width=\textwidth]{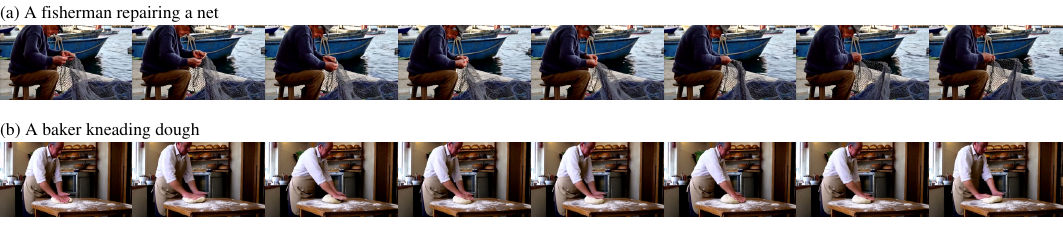}
\caption{\textbf{General text-to-video generation with Gravis-Video.} Text-only generations demonstrate the model's ability to synthesize everyday scenes and human--object interactions beyond robotic manipulation.}
\label{fig:appendix_general_t2v}
\end{minipage}\par\medskip

\subsection{Embodied Video Comparisons}
Figures~\ref{fig:appendix_robot_comparison} and~\ref{fig:appendix_ego_comparison} compare Gravis-Video with all eight evaluated baselines on two cases from the evaluation platform: bimanual knife manipulation and egocentric cloth folding.
Each comparison uses the common initial image and displays frames at matched elapsed times, preserving their original aspect ratios.
The robot case highlights gripper articulation, object geometry, and bimanual coordination, while the cloth case highlights hand appearance and deformable-object interaction.

\noindent\begin{minipage}[t]{\textwidth}
\captionsetup{type=figure}
\centering
\includegraphics[width=\textwidth]{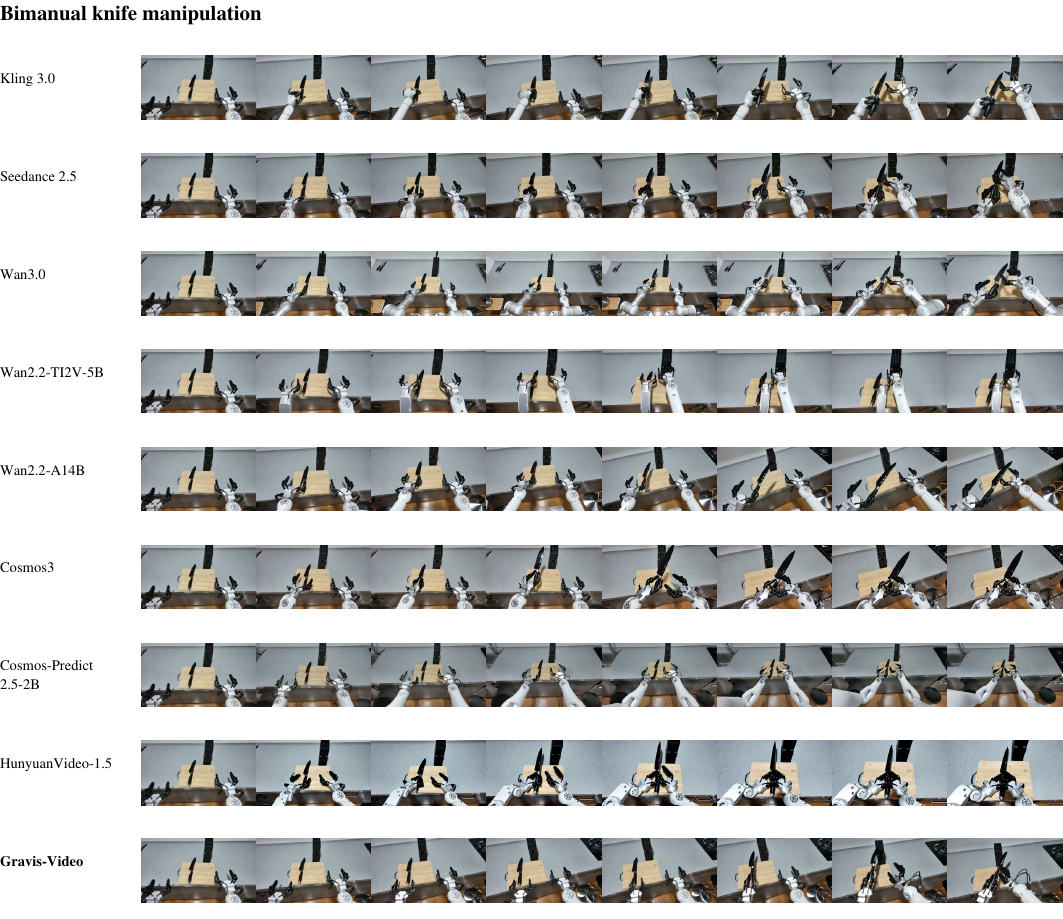}
\caption{\textbf{Bimanual manipulation across video generators.} Starting from the same image, each model generates a robot manipulating a knife with both grippers. Frames at matched elapsed times enable comparison of gripper articulation, object geometry, and bimanual coordination.}
\label{fig:appendix_robot_comparison}
\end{minipage}\par\medskip

\noindent\begin{minipage}[t]{\textwidth}
\captionsetup{type=figure}
\centering
\includegraphics[width=\textwidth]{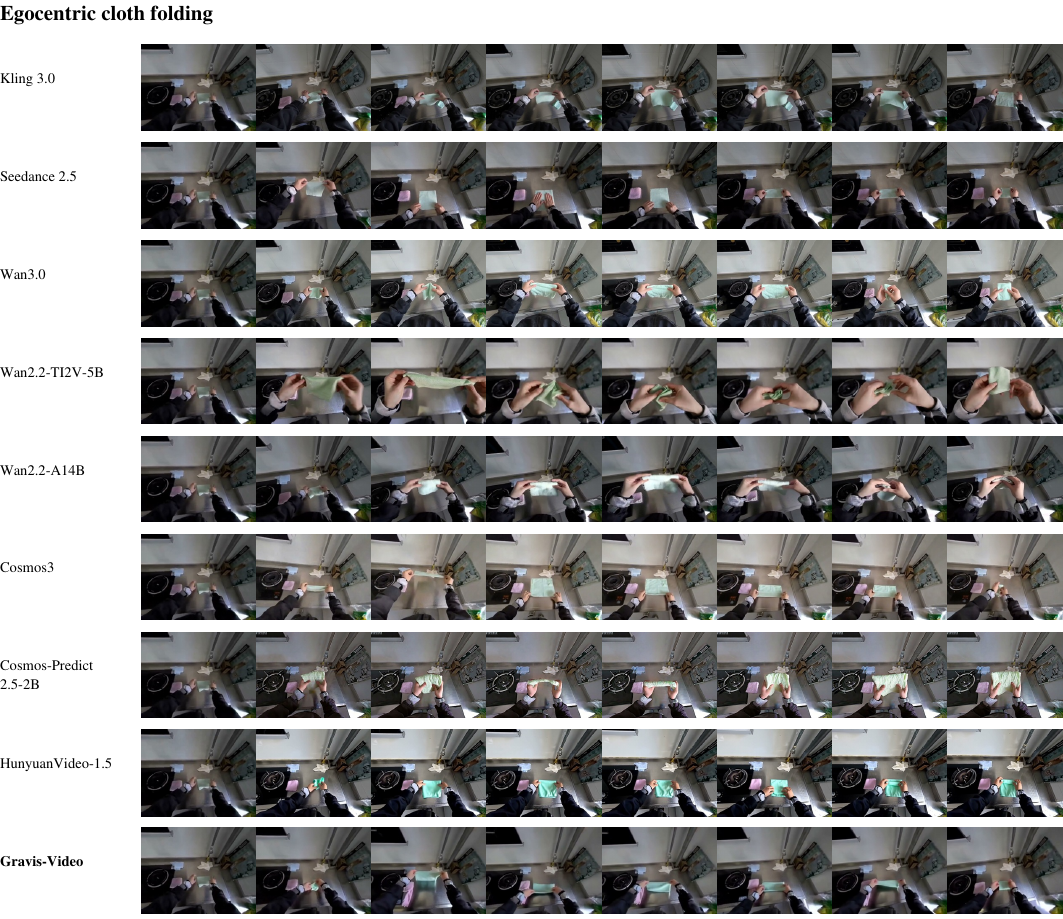}
\caption{\textbf{Egocentric manipulation across video generators.} Starting from the same image, each model generates a cloth-folding sequence. Temporally aligned frames highlight differences in hand structure, hand--object contact, and cloth deformation.}
\label{fig:appendix_ego_comparison}
\end{minipage}\par\medskip

\subsection{Direct Deployment after Embodied Pretraining}
Figure~\ref{fig:appendix_direct_rollouts} shows additional executions of the pretrained Gravis-WAM without task-specific post-training.
The instructions specify the acting arm, target object, and destination: placing a black object in the left container with the left arm, transferring a water bottle from the right hand to the left hand, and placing a mango on the tray with the right arm.
The first three sequences show instruction-guided manipulation among distractor objects; Headphone Hanging additionally shows the right hand picking up the headphones and placing them on their stand.

\noindent\begin{minipage}[t]{\textwidth}
\captionsetup{type=figure}
\centering
\includegraphics[width=.95\textwidth]{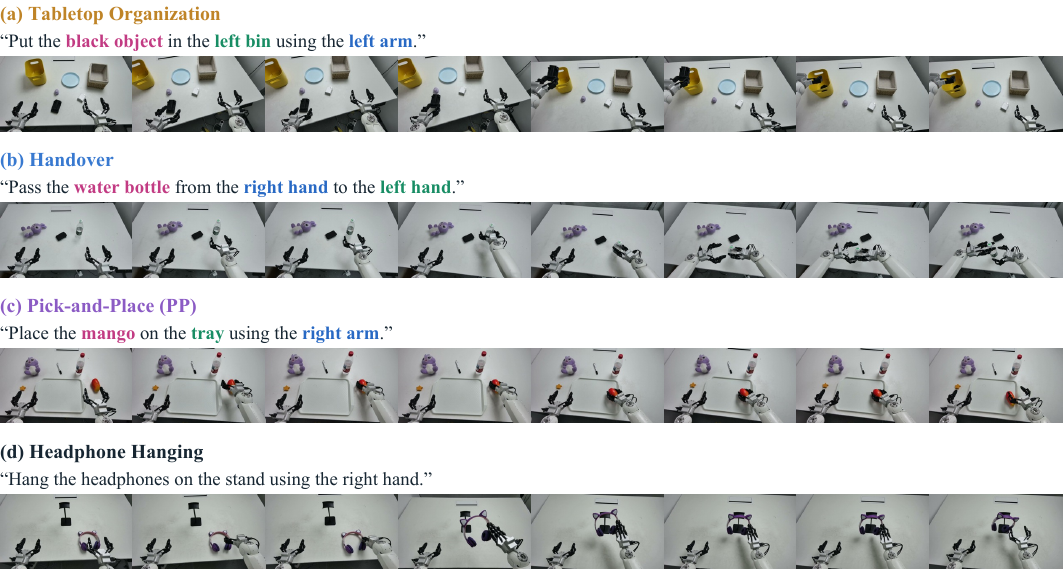}
\caption{\textbf{Direct deployment of pretrained Gravis-WAM.} Without task-specific post-training, the model performs Tabletop Organization, Handover, Pick-and-Place, and Headphone Hanging. The first three tasks require interpreting instructions specifying the acting arm, target object, and destination.}
\label{fig:appendix_direct_rollouts}
\end{minipage}\par\medskip

\subsection{Multi-Task Post-Training}
Figure~\ref{fig:appendix_post_rollouts} expands the successful executions shown in the main paper with eight key frames per task.
A single jointly adapted policy stows four chopsticks, hangs an electronic scale, and hangs two cups.

\noindent\begin{minipage}[t]{\textwidth}
\captionsetup{type=figure}
\centering
\includegraphics[width=.95\textwidth]{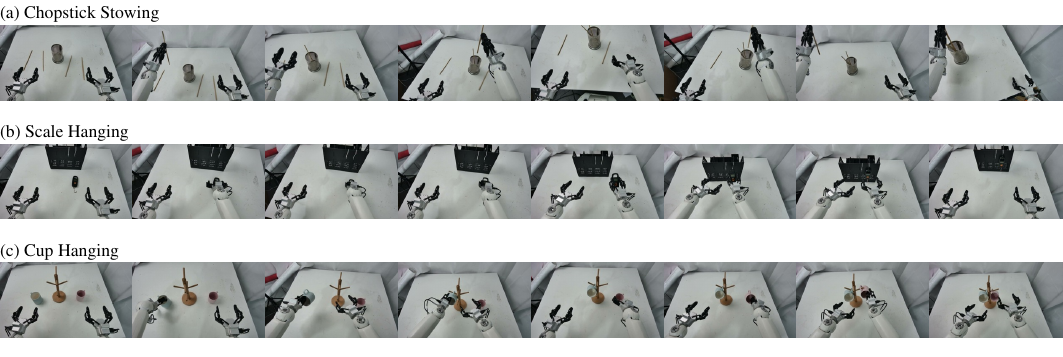}
\caption{\textbf{Real-world multi-task adaptation of Gravis-WAM.} A single jointly post-trained policy successfully performs Chopstick Stowing, Scale Hanging, and Cup Hanging.}
\label{fig:appendix_post_rollouts}
\end{minipage}\par\medskip

%% file: tables/robotwin_per_task_appendix.tex
% The original 50 task rows, displayed as two parallel groups of 25.
\begin{table}[!htbp]
\centering
\begingroup
\caption{\textbf{Per-task success rates on RoboTwin.} Each training protocol is evaluated in Clean and Random environments with 50 trials per task.}
\label{tab:robotwin_per_task}
\fontsize{8}{10}\selectfont
\setlength{\tabcolsep}{1.25pt}
\renewcommand{\arraystretch}{1.08}
\renewcommand{\tabularxcolumn}[1]{m{#1}}
\begin{tabularx}{\textwidth}{@{}>{\raggedright\arraybackslash}m{.205\textwidth}*{4}{>{\centering\arraybackslash}X}@{\hspace{12pt}}>{\raggedright\arraybackslash}m{.205\textwidth}*{4}{>{\centering\arraybackslash}X}@{}}
\toprule[.8pt]
\multirow{2}{*}{\textbf{Simulation task}} & \multicolumn{2}{c}{\textbf{Clean-only}} & \multicolumn{2}{c}{\textbf{Clean+Random}} & \multirow{2}{*}{\textbf{Simulation task}} & \multicolumn{2}{c}{\textbf{Clean-only}} & \multicolumn{2}{c}{\textbf{Clean+Random}}\\
& \textbf{C2C} & \textbf{C2R} & \textbf{Clean} & \textbf{Random} & & \textbf{C2C} & \textbf{C2R} & \textbf{Clean} & \textbf{Random}\\
\midrule
Adjust Bottle & 66\% & 24\% & 92\% & 96\% & Place Can Basket & 84\% & 42\% & 82\% & 82\% \\
Beat Block Hammer & 74\% & 36\% & 90\% & 86\% & Place Cans Plasticbox & 100\% & 82\% & 100\% & 100\% \\
Blocks Ranking RGB & 94\% & 70\% & 100\% & 100\% & Place Container Plate & 96\% & 100\% & 98\% & 98\% \\
Blocks Ranking Size & 66\% & 52\% & 92\% & 88\% & Place Dual Shoes & 68\% & 32\% & 96\% & 92\% \\
Click Alarmclock & 100\% & 84\% & 100\% & 100\% & Place Empty Cup & 100\% & 86\% & 98\% & 98\% \\
Click Bell & 100\% & 100\% & 100\% & 100\% & Place Fan & 88\% & 34\% & 98\% & 90\% \\
Dump Bin Bigbin & 90\% & 60\% & 100\% & 98\% & Place Mouse Pad & 88\% & 60\% & 96\% & 90\% \\
Grab Roller & 100\% & 92\% & 98\% & 100\% & Place Object Basket & 90\% & 62\% & 86\% & 82\% \\
Handover Block & 82\% & 34\% & 98\% & 98\% & Place Object Scale & 96\% & 68\% & 100\% & 98\% \\
Handover Mic & 86\% & 56\% & 98\% & 90\% & Place Object Stand & 100\% & 90\% & 98\% & 100\% \\
Hanging Mug & 38\% & 18\% & 62\% & 66\% & Place Phone Stand & 90\% & 34\% & 92\% & 100\% \\
Lift Pot & 100\% & 94\% & 100\% & 100\% & Place Shoe & 92\% & 74\% & 100\% & 100\% \\
Move Can Pot & 88\% & 46\% & 100\% & 96\% & Press Stapler & 80\% & 94\% & 94\% & 90\% \\
Move Pillbottle Pad & 86\% & 52\% & 96\% & 100\% & Put Bottles Dustbin & 90\% & 50\% & 92\% & 92\% \\
Move Playingcard Away & 96\% & 82\% & 100\% & 100\% & Put Object Cabinet & 50\% & 32\% & 92\% & 90\% \\
Move Stapler Pad & 84\% & 24\% & 92\% & 92\% & Rotate QR Code & 76\% & 22\% & 80\% & 66\% \\
Open Laptop & 88\% & 72\% & 98\% & 100\% & Scan Object & 66\% & 34\% & 88\% & 92\% \\
Open Microwave & 64\% & 64\% & 92\% & 98\% & Shake Bottle & 100\% & 90\% & 100\% & 100\% \\
Pick Diverse Bottles & 74\% & 66\% & 80\% & 86\% & Shake Bottle Horizontally & 100\% & 94\% & 100\% & 100\% \\
Pick Dual Bottles & 88\% & 74\% & 100\% & 96\% & Stack Blocks Three & 96\% & 62\% & 100\% & 96\% \\
Place A2B Left & 94\% & 72\% & 100\% & 100\% & Stack Blocks Two & 100\% & 76\% & 100\% & 100\% \\
Place A2B Right & 96\% & 46\% & 96\% & 94\% & Stack Bowls Three & 80\% & 78\% & 80\% & 88\% \\
Place Bread Basket & 90\% & 82\% & 90\% & 92\% & Stack Bowls Two & 100\% & 90\% & 100\% & 96\% \\
Place Bread Skillet & 92\% & 74\% & 96\% & 94\% & Stamp Seal & 98\% & 72\% & 100\% & 100\% \\
Place Burger Fries & 98\% & 78\% & 96\% & 92\% & Turn Switch & 92\% & 60\% & 94\% & 90\% \\
\midrule
\rowcolor{paperpale}\multicolumn{10}{c}{\textbf{Average (\%):}\quad C2C \textbf{87.08}\qquad C2R \textbf{63.40}\qquad Clean \textbf{94.60}\qquad Random \textbf{93.84}}\\
\bottomrule[.8pt]
\end{tabularx}
\endgroup
\end{table}

%% file: tables/ebench_per_task_appendix.tex
\begin{table}[!ht]
\centering\footnotesize
\caption{\textbf{Per-task success rates (\%) on EBench.} Following the official evaluation protocol, Gravis-WAM is evaluated over 510 trials: 15 each for Make Sandwich and Microwave, and 20 for every other task. Task Mean weights all 26 tasks equally.}
\label{tab:ebench_per_task}
\begingroup
\setlength{\tabcolsep}{2.25pt}
\renewcommand{\tabularxcolumn}[1]{m{#1}}
\renewcommand{\arraystretch}{1.0}
\begin{tabularx}{\textwidth}{@{}>{\raggedright\arraybackslash}m{.20\textwidth}*{8}{>{\centering\arraybackslash}X}@{}}
\toprule
Task & \shortstack{PICA-\\WAM} & \shortstack{OpenWAM-\\$\alpha$} & \shortstack{InternW0-\\$\Delta$} & \shortstack{GPT-6-\\Astra-ICL} & \shortstack{Qwen-\\RobotManip} & \shortstack{$\pi_{0.5}$} & \shortstack{GigaBrain-\\0.7} & \shortstack{\textbf{Gravis-}\\\textbf{WAM}}\\
\midrule
\rowcolor{paperpale}\multicolumn{9}{c}{\textit{Mobile Manipulation}}\\
Apple From Shelf & \textbf{80.00} & 70.00 & 15.00 & 70.00 & 65.00 & 75.00 & 55.00 & 70.00\\
Apple To Fruit Bowl & 95.00 & 95.00 & 95.00 & 90.00 & 50.00 & 25.00 & 20.00 & \textbf{100.00}\\
Bookmark On Book & 40.00 & 35.00 & 35.00 & \textbf{90.00} & 40.00 & 55.00 & 30.00 & 45.00\\
Bottle & \textbf{0.00} & \textbf{0.00} & \textbf{0.00} & \textbf{0.00} & \textbf{0.00} & \textbf{0.00} & \textbf{0.00} & \textbf{0.00}\\
Bowl To Plate & 75.00 & \textbf{85.00} & 80.00 & 75.00 & 75.00 & \textbf{85.00} & 65.00 & 45.00\\
Detergent & \textbf{100.00} & 95.00 & 95.00 & 55.00 & 85.00 & \textbf{100.00} & 95.00 & \textbf{100.00}\\
Dish & \textbf{50.00} & 40.00 & \textbf{50.00} & \textbf{50.00} & 20.00 & 25.00 & 5.00 & \textbf{50.00}\\
Dishwasher & \textbf{100.00} & 90.00 & 85.00 & 5.00 & 15.00 & 65.00 & 0.00 & 90.00\\
Fruit & \textbf{75.00} & 55.00 & 60.00 & 40.00 & 45.00 & 10.00 & 0.00 & 55.00\\
Make Sandwich & 66.67 & \textbf{80.00} & 73.33 & 46.67 & 66.67 & 66.67 & \textbf{80.00} & 40.00\\
Microwave & 80.00 & \textbf{93.33} & 6.67 & 73.33 & 33.33 & 40.00 & 0.00 & 80.00\\
Pen & 10.00 & 5.00 & \textbf{15.00} & \textbf{15.00} & 5.00 & 0.00 & 0.00 & 10.00\\
Perfume To Cosmetics Rack & 60.00 & 75.00 & 30.00 & 75.00 & 55.00 & 60.00 & 25.00 & \textbf{85.00}\\
Remote To Holder & 85.00 & 65.00 & 85.00 & \textbf{100.00} & 65.00 & 60.00 & 50.00 & 95.00\\
Salt To Spice Rack & 95.00 & \textbf{100.00} & \textbf{100.00} & \textbf{100.00} & 70.00 & 70.00 & 75.00 & \textbf{100.00}\\
Shop & \textbf{100.00} & 0.00 & 60.00 & 0.00 & 0.00 & 0.00 & 0.00 & 85.00\\
Soap To Dish & 95.00 & 85.00 & \textbf{100.00} & 95.00 & 85.00 & \textbf{100.00} & 70.00 & 90.00\\
Teacup To Saucer Teapot To Tray & 45.00 & 40.00 & 45.00 & \textbf{85.00} & 45.00 & 30.00 & 20.00 & 65.00\\
Utensils To Holder & 30.00 & \textbf{35.00} & 15.00 & 10.00 & 15.00 & 5.00 & 10.00 & 20.00\\
\rowcolor{paperpale}\multicolumn{9}{c}{\textit{Tabletop Manipulation}}\\
Collect Coffee Beans & 20.00 & 20.00 & 5.00 & 0.00 & 5.00 & 30.00 & \textbf{50.00} & 20.00\\
Flip Cup Collect Cookies & 40.00 & 40.00 & 50.00 & 10.00 & \textbf{90.00} & 25.00 & 40.00 & 75.00\\
Frame Against Pen Holder & 70.00 & 50.00 & 60.00 & 65.00 & \textbf{95.00} & 45.00 & 70.00 & 85.00\\
Install Gear & \textbf{55.00} & 40.00 & 50.00 & 15.00 & 50.00 & 10.00 & 20.00 & 30.00\\
Peg In Hole & 30.00 & 40.00 & 10.00 & 20.00 & 10.00 & 10.00 & 15.00 & \textbf{45.00}\\
Put Glass In Glassbox & 60.00 & 75.00 & 60.00 & 20.00 & \textbf{90.00} & 65.00 & 55.00 & 85.00\\
Tighten Nut & 40.00 & 30.00 & 0.00 & 10.00 & 10.00 & 20.00 & 15.00 & \textbf{60.00}\\
\midrule
\rowcolor{paperpale}\textbf{Task Mean} & 61.41 & 55.32 & 49.23 & 46.73 & 45.58 & 41.41 & 33.27 & \textbf{62.50}\\
\bottomrule
\end{tabularx}
\endgroup
\end{table}

\begin{table}[!ht]
\centering\footnotesize
\caption{\textbf{Per-task stage-completion scores on EBench.} Scores range from 0 to 100 and credit partial task completion. Evaluation counts follow Table~\ref{tab:ebench_per_task}; Task Mean weights all 26 tasks equally.}
\label{tab:ebench_per_task_score}
\begingroup
\setlength{\tabcolsep}{2.25pt}
\renewcommand{\tabularxcolumn}[1]{m{#1}}
\renewcommand{\arraystretch}{1.0}
\begin{tabularx}{\textwidth}{@{}>{\raggedright\arraybackslash}m{.20\textwidth}*{8}{>{\centering\arraybackslash}X}@{}}
\toprule
Task & \shortstack{PICA-\\WAM} & \shortstack{OpenWAM-\\$\alpha$} & \shortstack{InternW0-\\$\Delta$} & \shortstack{GPT-6-\\Astra-ICL} & \shortstack{Qwen-\\RobotManip} & \shortstack{$\pi_{0.5}$} & \shortstack{GigaBrain-\\0.7} & \shortstack{\textbf{Gravis-}\\\textbf{WAM}}\\
\midrule
\rowcolor{paperpale}\multicolumn{9}{c}{\textit{Mobile Manipulation}}\\
Apple From Shelf & \textbf{80.00} & 70.00 & 15.00 & 70.00 & 65.00 & 75.00 & 55.00 & 70.00\\
Apple To Fruit Bowl & 95.00 & 95.00 & 95.00 & 90.00 & 50.00 & 25.00 & 20.00 & \textbf{100.00}\\
Bookmark On Book & 40.00 & 35.00 & 35.00 & \textbf{90.00} & 40.00 & 55.00 & 30.00 & 45.00\\
Bottle & 43.00 & 32.00 & 44.00 & 17.00 & 17.00 & 5.00 & 4.00 & \textbf{49.57}\\
Bowl To Plate & 75.00 & \textbf{85.00} & 80.00 & 75.00 & 75.00 & \textbf{85.00} & 65.00 & 45.00\\
Detergent & \textbf{100.00} & 98.00 & 98.00 & 80.00 & 93.00 & \textbf{100.00} & 98.00 & \textbf{100.00}\\
Dish & \textbf{90.00} & 88.00 & 89.00 & 81.00 & 76.00 & 70.00 & 30.00 & \textbf{90.00}\\
Dishwasher & \textbf{100.00} & 95.00 & 93.00 & 53.00 & 27.00 & 73.00 & 50.00 & 92.50\\
Fruit & \textbf{88.00} & 76.00 & 84.00 & 59.00 & 68.00 & 38.00 & 27.00 & 84.00\\
Make Sandwich & 92.00 & \textbf{95.00} & 93.00 & 70.00 & 90.00 & 92.00 & \textbf{95.00} & 73.33\\
Microwave & 96.00 & \textbf{98.00} & 64.00 & 90.00 & 38.00 & 50.00 & 0.00 & 92.22\\
Pen & 42.00 & 37.00 & 32.00 & \textbf{55.00} & 25.00 & 5.00 & 3.00 & 48.33\\
Perfume To Cosmetics Rack & 63.00 & 78.00 & 30.00 & \textbf{88.00} & 60.00 & 68.00 & 33.00 & 87.50\\
Remote To Holder & 85.00 & 65.00 & 85.00 & \textbf{100.00} & 65.00 & 60.00 & 50.00 & 95.00\\
Salt To Spice Rack & 95.00 & \textbf{100.00} & \textbf{100.00} & \textbf{100.00} & 70.00 & 70.00 & 75.00 & \textbf{100.00}\\
Shop & \textbf{100.00} & 67.00 & 90.00 & 60.00 & 60.00 & 58.00 & 30.00 & 96.25\\
Soap To Dish & 95.00 & 85.00 & \textbf{100.00} & 95.00 & 85.00 & \textbf{100.00} & 70.00 & 90.00\\
Teacup To Saucer Teapot To Tray & 58.00 & 53.00 & 58.00 & \textbf{93.00} & 50.00 & 40.00 & 28.00 & 75.00\\
Utensils To Holder & 50.00 & \textbf{55.00} & 45.00 & 23.00 & 40.00 & 30.00 & 23.00 & 47.50\\
\rowcolor{paperpale}\multicolumn{9}{c}{\textit{Tabletop Manipulation}}\\
Collect Coffee Beans & 50.00 & 40.00 & 11.00 & 24.00 & 51.00 & 55.00 & 71.00 & \textbf{74.29}\\
Flip Cup Collect Cookies & 84.00 & 59.00 & 91.00 & 44.00 & \textbf{97.00} & 43.00 & 61.00 & 87.00\\
Frame Against Pen Holder & 77.00 & 60.00 & 70.00 & 80.00 & \textbf{97.00} & 48.00 & 72.00 & 85.00\\
Install Gear & \textbf{65.00} & 50.00 & 60.00 & 20.00 & 58.00 & 13.00 & 28.00 & 37.50\\
Peg In Hole & 48.00 & \textbf{60.00} & 33.00 & \textbf{60.00} & 35.00 & 23.00 & 55.00 & 50.00\\
Put Glass In Glassbox & 70.00 & 85.00 & 72.00 & 30.00 & \textbf{95.00} & 77.00 & 68.00 & 85.00\\
Tighten Nut & 70.00 & 63.00 & 50.00 & 55.00 & 55.00 & 58.00 & 58.00 & \textbf{80.00}\\
\midrule
\rowcolor{paperpale}\textbf{Task Mean} & 75.04 & 70.15 & 66.04 & 65.46 & 60.85 & 54.46 & 46.12 & \textbf{76.15}\\
\bottomrule
\end{tabularx}
\endgroup
\end{table}